\documentclass[fleqn]{sig-alternate}

\usepackage[
  pass,% keep layout unchanged
]{geometry}

\usepackage{bbm}
\usepackage{hyperref}
\usepackage{caption}
\usepackage{color}
\usepackage{xspace}
\usepackage[utf8x]{inputenc}
\usepackage{graphicx}
\usepackage[lined, ruled, boxed, commentsnumbered]{algorithm2e}
\usepackage{amsmath,amssymb}
\usepackage{booktabs}
\usepackage{verbatim}
\usepackage{subcaption}
\DeclareCaptionType{copyrightbox}
\usepackage{subcaption}
\usepackage{booktabs}
\usepackage{times}
\usepackage{microtype}
\usepackage{color}

\newcommand*{\myindent}{\hspace*{0.4cm}}%

\newtheorem{exm}{Example}[section]

\newcommand{\example}[1]{\vspace*{-0.7em}\begin{exm}\small{#1}\end{exm}\vspace*{-0.7em}}
\begin{document}

\numberofauthors{1} %  in this sample file, there are a *total*
\newcommand{\authspace}{\hspace*{.2in}}
\author{Subhabrata Mukherjee \authspace \authspace \authspace Gerhard Weikum \\
 \affaddr{Max Planck Institute for Informatics} \\
 \email{\{smukherjee, weikum\}}@mpi-inf.mpg.de
 }

%%%subho: need a better title
%People on Media: \\ Credibility and Polarization of Users in News Communities
%Credibility-aware News Recommendation
%Interaction on Media: Credibility of Information and Interactions in News Communities
\title{Leveraging Joint Interactions for \\ Credibility Analysis in News Communities
}
%%%GW: why polarization? what is polarization exactly? are we using this term later on, explain it, analyze it at all?

\newcommand{\squishlist}{
   \begin{list}{$\bullet$}
    { \setlength{\itemsep}{0pt}      \setlength{\parsep}{3pt}
      \setlength{\topsep}{3pt}       \setlength{\partopsep}{0pt}
      \setlength{\leftmargin}{1.5em} \setlength{\labelwidth}{1em}
      \setlength{\labelsep}{0.5em} } }
\newcommand{\squishlisttwo}{
   \begin{list}{$\bullet$}
    { \setlength{\itemsep}{0pt}    \setlength{\parsep}{0pt}
      \setlength{\topsep}{0pt}     \setlength{\partopsep}{0pt}
      \setlength{\leftmargin}{0.5em} \setlength{\labelwidth}{0.5em}
      \setlength{\labelsep}{0.5em} } }

\newcommand{\squishend}{
    \end{list} 
}

\newcommand{\comm}[1]{}

\maketitle

\section*{Abstract}
%%%GW: trimmed abstract to make it crisper

%Motivation
Media seems to have become more partisan, often providing a biased coverage of news catering to the interest of specific groups. It is therefore essential to identify {\em credible} information content that provides an objective narrative of an event.
%Why is it difficult
News communities such as digg, reddit, or newstrust offer recommendations, reviews, quality ratings, and 
further insights on journalistic works.
However, there is a complex {\em interaction} between different factors in such online communities:
fairness and style of reporting, 
language clarity and objectivity, 
topical perspectives (like political viewpoint), 
expertise and bias of community members, and more.
%Furthermore, the ground-truth required for such a kind of study is difficult to obtain as the human judgment is also influenced by their %subjectivity like perspectives, topic preference, expertise and bias.

%Approach and Contribution
This paper presents a model to systematically analyze the different interactions in a news community between users, news, and sources. We develop a probabilistic graphical model that leverages this {\em joint} interaction to identify 1) highly {\em credible} news articles, 2) {\em trustworthy} news sources, and 3) {\em expert} users who perform the role of ``citizen journalists'' in the community.
Our method extends CRF models to incorporate real-valued ratings, as some communities have very fine-grained scales 
that cannot be easily discretized without losing information.
To the best of our knowledge, this paper is the first full-fledged analysis of credibility, trust, and expertise in news communities.
%%%GW: are these the three key dimensions: credibility, trust, expertise ?!
%%%subho: yes
%To the best of our knowledge, this is the first full-fledged study on a news community dataset that analyzes the different kinds of %interactions therein for credibility analysis.

%Media is widely regarded as the {fourth} pillar of democracy, tasked with providing people with credible information in a meaningful context for decision-making.  In this work, we analyze the influence of different facets like \textit{topics, linguistic features, viewpoint} and \textit{format} on the credibility of news from media (newspapers, tv, blogs, online sources etc.) as viewed and expressed by different users in a \textit{news community}. We analyze  different kinds of interactions taking place between users, media and news. We leverage the interplay between different factors in a probabilistic graphical model based approach
%Since credibility is subjective, the work attempts to find the different factors that polarize a community from the point of view of different users.

\vspace{-1mm}
\ -\ \textit{Information Filtering}
%A category including the fourth, optional field follows...
\ -\ \textit{Natural Language Processing}
\vspace*{-1.5mm}
%\terms{Design, Algorithms, Measurement, Experimentation}
\vspace*{-1.5mm}
 % NOT required for Proceedings
\vspace*{-0.5mm}

\section{Introduction}
\label{sec:intro}

\noindent{\bf Motivation:}
Media plays a crucial role in the public dissemination of information about events. %However 70 percent of Americans believe that there is either a great deal or a fair %amount of media bias in news coverage~\cite{nber}. A June 2013 Gallup %poll~\cite{gallop,global-research} indicates that nearly 4 out of 5 Americans among %younger generations from age 21-64 do not trust the major news networks in the age %of super-mergers; when corporations like General Electric, Comcast and possibly Time %Warner own media in the likes of NBC and MSNBC.
%%%GW: need to make wording much crisper
Many people find online information and blogs as useful as TV or magazines.
%%%GW: no citation needed here
At the same time, however, people also believe that there is substantial media bias in
news coverage \cite{nber,gallop}, %cite{nber}
%%%GW: we should reconsider if we really need 3 citations here
especially in view of inter-dependencies and cross-ownerships
of media companies and other industries (like energy).

Several factors affect the coverage and presentation 
of news in media incorporating potentially biased information induced via the fairness and style of reporting.
%%%GW: isn't this trivial: Politics always comes with political orientation,
%%% so political news are naturally more susceptible to bias than say sports
%One such factor is the \emph{topical coverage} of news. 
%We found that $54\%$ of the news coverage on any topic (and $62\%$ of the news %articles) is related to \emph{Politics} {explicitly}, which, in turn, is influenced by the %political orientation of the media (e.g., left, center, or right). The political viewpoint %and perspective of the users and news sources (e.g., BBC, CNN) lead to polarization in %the community. The influence of Politics in media~\cite{chomsky1988} and vice %versa~\cite{dellavigna2007} has received vigorous interest in recent times.
%%%subho: influence of Politics on even sports (election) or environment (e.g., global warming)
%GW: keep the intro straight, don't add digression on secondary issues
%For instance, we found significant influence of {\em Politics} on almost all topics covered by the media.
News 
%on such topics 
are often presented in a polarized way 
depending on the political viewpoint of the media source (newspapers, TV stations, etc.). 
In addition, other  source-specific properties like {\em viewpoint, expertise, and format} of news may also be indicators of information credibility. 
%\cite{flanagin2000} found that people consider online information to be as credible as that obtained from television, radio, and magazines, %but not as credible as newspaper information. \cite{johnson2007} reported that politically interested Internet users find blogs to be %moderately credible sources for news and other information.

%\noindent{\bf News Community:} 
In this paper, we embark on an in-depth study and formal modeling 
of these factors and
inter-dependencies within \textit{news communities} for {\em credibility analysis}. A news community is a news aggregator site (e.g., reddit.com, digg.com, newstrust.net) where users can give explicit feedback (e.g., rate, review, share) on the quality of news and
can interact (e.g., comment, vote) with each other. Users can rate and review news, point out differences, bias in perspectives, unverified claims etc. However, 
this adds user subjectivity to the evaluation process, as users incorporate their own bias and perspectives in the framework. Controversial topics create polarization among users which influence their ratings. \cite{sloanreview,fang2014} state that online ratings are one of the most trusted sources of user feedback; however they are systematically {\em biased} and easily manipulated.

%\subsection{Research Questions}
%{\color{red}
%\noindent{\bf Research Questions:}
%\input{objectives}
%}

{
\noindent{\bf Problem Statement:} Given a set of news sources generating news articles, and users reviewing those articles on different qualitative aspects with mutual interactions --- our objective is to {\em jointly} rank the sources, articles, and users based on their trustworthiness, credibility, and expertise respectively.

In this process, we want to analyze the influence of various factors like the writing style of a news article, its topic distribution, type of media and format of news, political viewpoint and expertise, and other user traits on the {\em credibility analysis} of the community.
}

\noindent{\bf Our Approach:} 
%%%GW: claiming the study of this data as a contribution was adequate for ICSWM 
%%%but not for CIKM - here emphasis must be on methodology
%In this work we present an in-depth study of the different kinds of interactions at play %in a typical news community that influences the credibility of the information content %therein. One of our contribution is in harvesting a rich (and hitherto untapped) %dataset to analyze these interactions, consisting of $~62$K news articles from $~6$K %news sources that are rated and analyzed on different facets by $~6$K expert users %contributing $~134$K reviews in {\em newstrust.net}.
%
To analyze the factors and inter-dependencies in a news community,
we have developed a sophisticated probabilistic graphical model, 
specifically a Continuous Conditional Random Field (CCRF) model, 
which exploits several {\em moderate} signals of interaction {\em jointly} between the following factors to 
derive a {\em strong} signal for information credibility {(refer to Figures~\ref{subfig1:interaction} and \ref{subfig2:instance})}. In particular, the model captures the following factors.

\squishlisttwo
\item \textit{Language and credibility of a news article}: \textit{objectivity}, rationality, and general quality of language in the news article. Objectivity is the quality of the news to be free from emotion, bias and prejudice of the author. The \textit{credibility} of a news article refers to presenting an unbiased, informative and balanced narrative of an event.
\item \textit{Properties and trustworthiness of a news source}: \textit{trustworthiness} of a news source in the sense of generating credible articles based on 
source properties like viewpoint, expertise 
%type
%%%subho: type - newspaper, radio, blog; format - editorial, investigative, research
%%%GW: what is "type"? whay is it not subsumed under "format"?
and format of news.
\item \textit{Expertise of users and review ratings}: \textit{expertise} of a user,
in the news community, in properly judging the credibility of news articles. 
Expert users should provide objective evaluations -- by reviews and/or ratings -- of news articles, corroborating with the evaluations of other expert users. This can be used to identify potential ``citizen journalists'' \cite{lewis2010} in the community.
\squishend

We show that the CCRF performs better than sophisticated collaborative filtering approaches based on latent factor models, and regression methods that do not consider all these interactions. 

{Although this work is focused on news communities, the framework can also be used for instance, in health communities (e.g. {\small \tt healthboards.com}) where users write posts on drug usage --- the objective being to {\em jointly} rank posts, drug side-effects, and users based on their quality, credibility, and trustworthiness respectively.
}

In this work, the attributes {\em credibility} and {\em trustworthiness} are
always associated with a news article and a news source, respectively. The joint
interaction between several factors also captures that a source garners
trustworthiness by generating credible news articles, which are highly rated by
expert users. Similarly, the likelihood of a news article being credible
increases if it is generated by a trustworthy source.

%%%GW: important to state the following, as we claim that the Continuous CRF is a specific contribution
Some communities offer users {\em fine-grained scales for rating} different
aspects of news articles and news sources. For example, the {\em newstrust.net} community analyzes an article on $15$ aspects like insightful, fairness, style and factual. These are aggregated into an overall {\em real-valued} rating after weighing the aspects based on their importance, expertise of the user, feedback from the community, and more.
%Newstrust, for example, has aspects like {\tt GW: .......... ?????  (fill in)}, each of which can be rated in 0.1 steps between 0 and 5, and can further be aggregated into an overall rating.
This setting cannot be easily discretized without
blow-up or risking to lose information. Therefore, we model ratings as
real-valued variables in our CCRF.

\noindent{\bf Contributions:}
%%%GW: now a few sentences on contributions, so far was more on approach
The paper introduces the following novel elements:

\squishlisttwo
\item A continuous CRF that captures the mutual dependencies between
credibility of articles, trustworthiness of sources, expertise of users,
and expresses real-valued ratings.
\item An inference method for the CCRF that allows us to {\em jointly} (a) predict ratings;
and (b) rank articles, sources, and users by their credibility, trustworthiness,
and expertise, respectively.
\item A large experimental study with data from {\em newstrust.net},
one of the most sophisticated news communities with a focus on quality journalism.
\squishend

The rest of the paper is organized as follows.
Section 2 presents how we model news communities, and which
factors we include in the model.
Section 3 develops the CCRF that captures the interaction between all the factors.
Section 4 introduces the dataset that we use for experimental evaluation
and further studies.
Section 5 presents our experimental results followed by discussion.

\section{Modeling News Communities}
\label{sec:newscom}
%In this section, we study the different factors of a news community and their interplay that %influence the credibility of news.

%\subsection{Overview of the Model}
%\label{subsec:overview}

%%%GW: make the figure 2-column wide (again) - right now it looks cramped
\begin{figure}
\begin{subfigure}[b]{0.5\textwidth}
        \centering
	\includegraphics[width=0.7\textwidth, height=50mm]{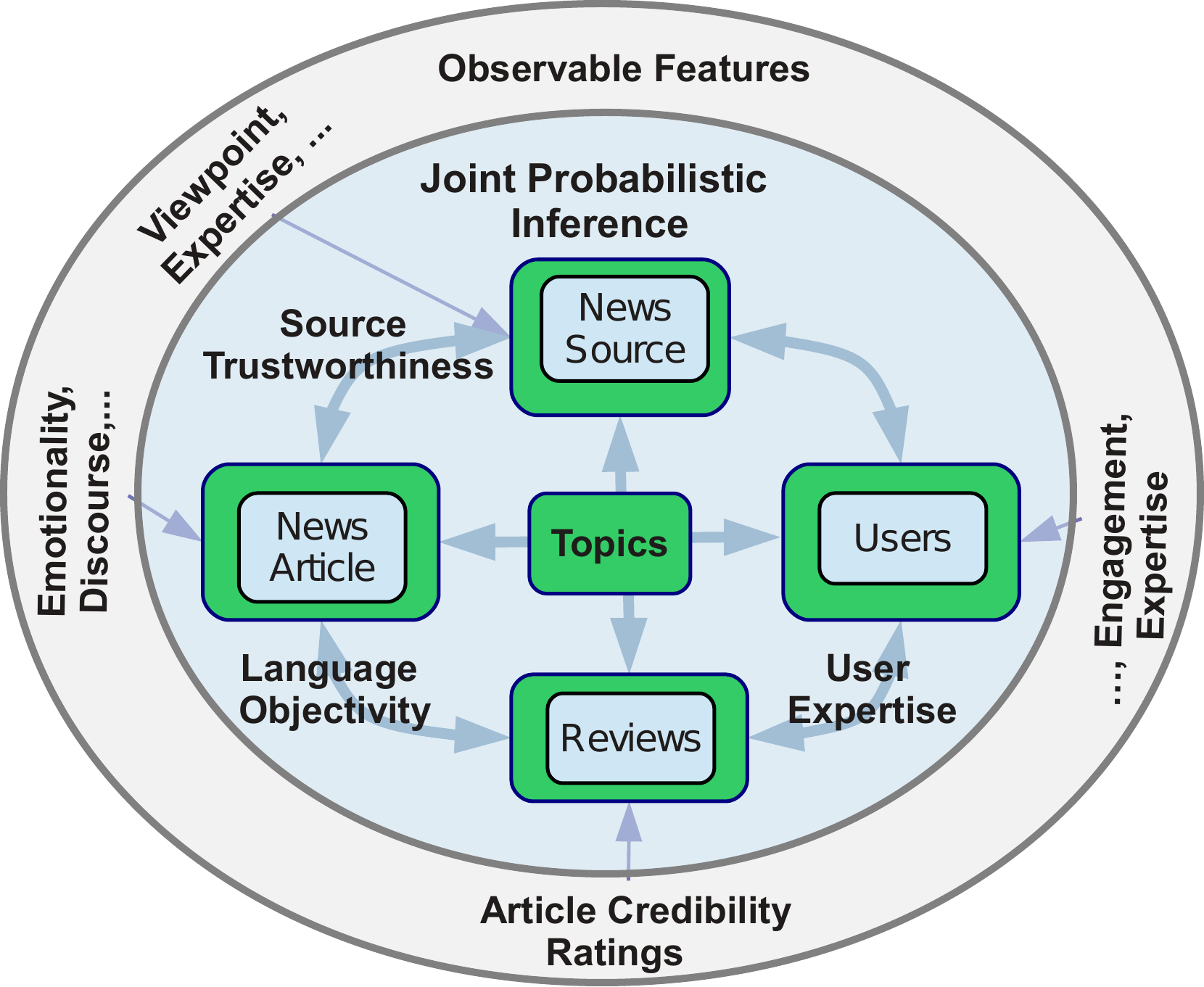}
	\caption{Interactions between source trustworthiness, article credibility, language objectivity, and user expertise.}
	\label{subfig1:interaction}
	\vspace{-1em}
    \end{subfigure}
    \hfill
    \begin{subfigure}[b]{0.45\linewidth}
        \centering
	\includegraphics[width=0.9\linewidth, height=35mm]{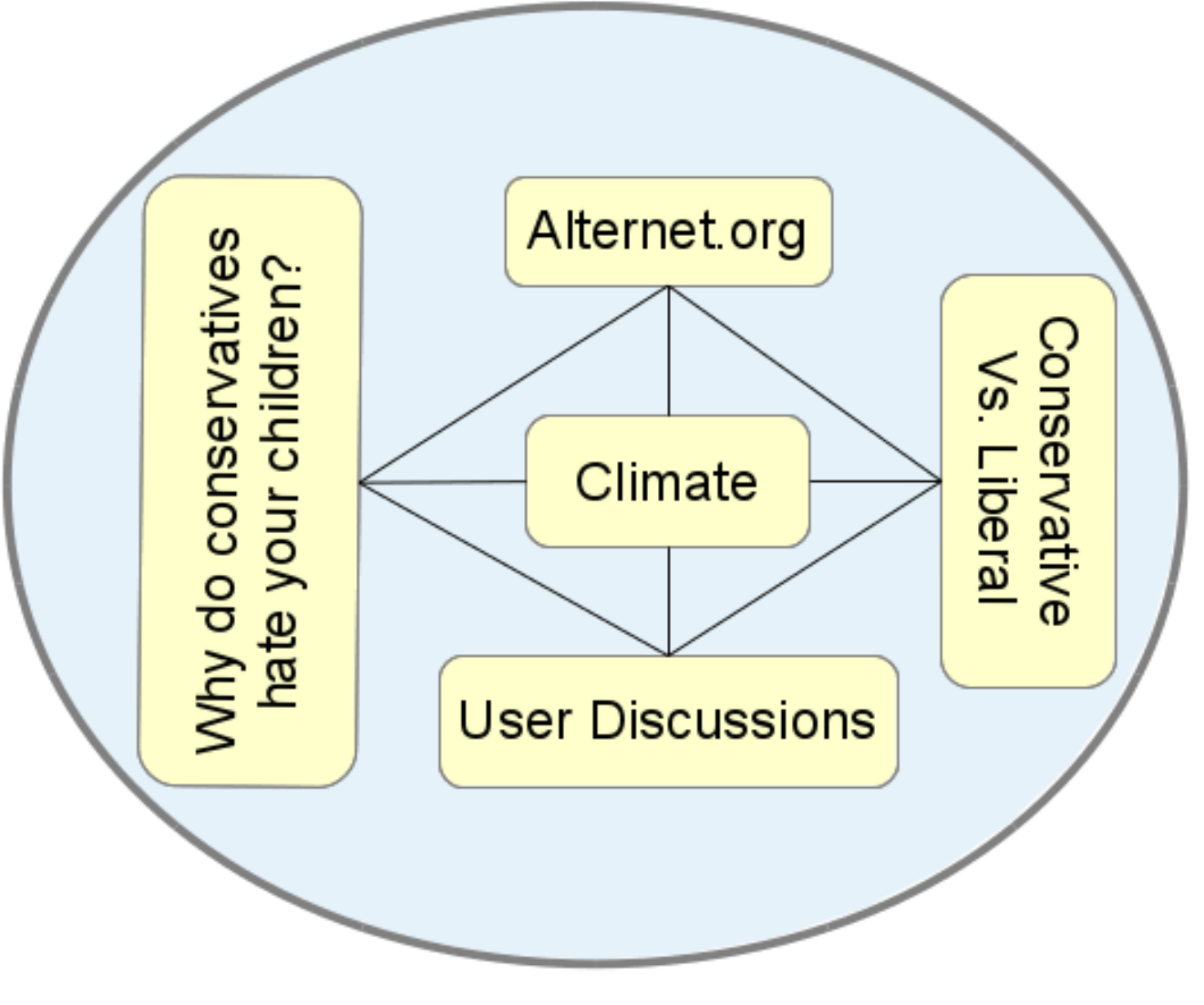}
	\caption{Sample instantiation.}
	\label{subfig2:instance}
    \end{subfigure}
        \begin{subfigure}[b]{0.6\linewidth}
        \centering
	\includegraphics[ height=50mm]{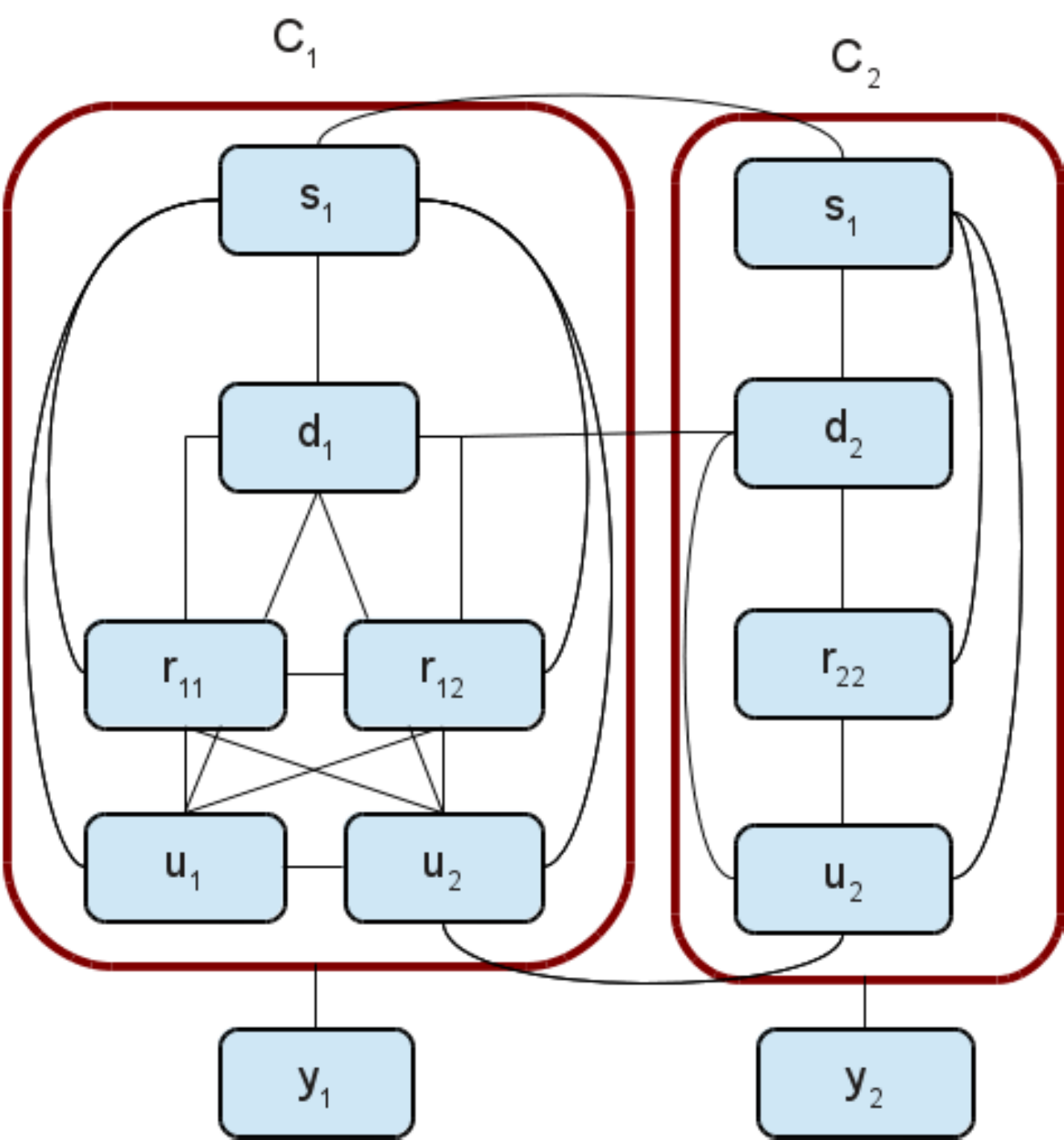}
	\caption{Clique representation.}
	\label{subfig3:clique}
    \end{subfigure}
    \vspace{-2em}
    \caption{Graphical model representation.}
    \label{fig:0}
 \vspace{-2.2em}
\end{figure}

%%%GW: included model overview in the modeling section
%{\tt GW: the whole text here needs more polishing - sometimes it is very compact and hard to follow, sometimes it is pretty verbose}\\

%\label{subfig1:interaction},\label{subfig1:clique}, \label{subfig1:instantiation} 
Our approach exploits the rich interaction taking place between the different factors in a news community. We propose a {\em probabilistic graphical model} that leverages the interplay between news credibility, language objectivity, source trustworthiness, and user expertise. Refer to Figure~\ref{fig:0} for the following discussion.

Consider a set of news sources $\langle s \rangle$ (e.g., $s_1$ in Figure~\ref{subfig3:clique}) generating articles $\langle d \rangle$ which are reviewed and analyzed by users $\langle u \rangle$ for their credibility. Consider $r_{ij}$ to be the review by user $u_j$ on article $d_i$. The overall article rating of $d_i$ is given by $y_i$.

%In our model, each news source (e.g., $s_1$ in Figure~\ref{fig:0}), news article (e.g., $a_1$), user (e.g., $u_1$) and his rating or review (e.g., $r_11$) is associated with a continuous random variable $\in [1 - 5]$, that indicates its trustworthiness, objectivity, expertise and credibility respectively. $5$ denotes the maximum quality that can be achieved by any node.

In our model, each news source, news article, user and her rating or review, and overall article rating is associated with a continuous random variable %{\color{red}$y$} 
{$r.v.\in [1 \ldots 5]$}, that indicates its trustworthiness, objectivity, expertise, and credibility, respectively. $5$ indicates the best quality that an item can obtain, and $1$ is the worst. {Discrete ratings, being a special case of this setting, can be easily handled.}

Each node is associated with a set of observed features that are extracted from the news community. For example, a news source has properties like topic specific expertise, viewpoint and format of news; a news article has features like topics, and style of writing from the usage of discourse markers and subjective words in the article. For users we extract their topical perspectives and expertise, engagement features (like the number of questions, replies, reviews posted) and various interactions with other users (like upvotes/downvotes) and news sources in the community.

The objective of our  model is to 
%retrieve 
predict
credibility ratings $\langle y \rangle$ of news articles $\langle d \rangle$ by exploiting the mutual interactions between different variables. The following edges between the variables capture their interplay:
\squishlisttwo
\item Each news article is connected to the news source from where it is extracted (e.g., $s_1-d_1$, $s_1-d_2$)
\item Each news article is connected to its review or rating by a user (e.g., $d_1-r_{11}$, $d_1-r_{12}$, $d_2-r_{22}$)
\item Each user is connected to all her reviews (e.g., $u_1-r_{11}$, $u_2-r_{12}$, $u_2-r_{22}$)
\item Each user is connected to all news articles rated by her (e.g., $u_1-d_1$, $u_2-d_1$, $u_2-d_2$)
\item Each source is connected to all the users who rated its articles (e.g., $s_1-u_1$, $s_1-u_2$)
\item Each source is connected to all the reviews of its articles (e.g., $s_1-r_{11}$, $s_1-r_{12}$, $s_1-r_{22}$)
\item For each article, all the users and all their reviews on the article are inter-connected (e.g., $u_1-r_{12}$, $u_2-r_{11}$, $u_1-u_2$). This captures user-user interactions (e.g., $u_1$ upvoting/downvoting $u_2$'s rating on $d_1$) influencing the overall article rating.
\squishend

Therefore, a {\em clique} (e.g., $C_1$) is formed between a news article, its source, users and their reviews on the article. Multiple such cliques (e.g., $C_1$ and $C_2$) share information via their common news sources (e.g., $s_1$) and users (e.g., $u_2$).

News \textit{topics} play a significant role on information credibility. Individual users in community (and news sources) have their own perspectives and expertise on various topics (e.g., environmental politics). Modeling user-specific topical perspectives explicitly captures credibility judgment better than a user-independent model.
However, many articles do not have explicit topic tags. 
%%%GW: correct? LDA is latent topics, not explicit - here, too?
%%%subho: yes
Hence we use Latent Dirichlet Allocation (LDA)~\cite{blei2003} in conjunction with Support Vector Regression (SVR)~\cite{drucker1996} to learn words associated to each (latent) topic, and user (and source) perspectives for the topics. Documents are assumed to have a distribution over
%Latent Dirichlet Allocation assumes a document to have a distribution over topics, and topics to have a distribution over words.
topics as latent variables, with words as observables. Inference is by Gibbs sampling.
This LDA model is a component of the overall model, discussed next.
%and hyper-parameter optimization is done by Support Vector Regression to guide the sampling process by attributing more mass to the topics for which the users have a higher preference.

We use a probabilistic graphical model, specifically a Conditional Random Field (CRF), to model all factors jointly. 
The modeling approach is related to
the prior work of~\cite{mukherjee2014}.
%%%GW: toned down the wording here: "related to" 
However, unlike that work and traditional CRF models, our problem setting requires a \textit{continuous} version of the CRF (CCRF) to deal with real-valued ratings instead of discrete labels. In this work, we follow an approach similar to~\cite{qinNIPS2008,radosavljevicECAI2010,tadas14} in learning the parameters of the CCRF. We use Support Vector Regression~\cite{drucker1996} to learn the elements of the feature vector for the CCRF.

The inference is centered around cliques of the form
 $\langle$ source, article, $\langle$ users $\rangle$, $\langle$ reviews $\rangle \rangle$.
An example is the two cliques $C_1: ~ s_1-d_1-\langle u_1, u_2 \rangle - \langle r_{11}, r_{12} \rangle$
and $C_2: ~ s_1-d_2-u_2-r_{22}$ in the instance graph of Figure \ref{subfig3:clique}.
This captures the ``cross-talk'' between different cliques sharing nodes. 
A news source garners trustworthiness by generating multiple credible articles. Users attain expertise by correctly identifying credible articles that corroborate with other expert users. Inability to do so brings down their expertise. Similarly, an article attains credibility if it is generated by a trustworthy source and highly rated by an expert user.
The inference algorithm for the CCRF is discussed in detail in Section \ref{sec:inference}.

%%%GW: added explicit structure
In the following subsections, 
we discuss the various feature groups that are considered in
our CCRF model.

\vspace{-0.2em}
%\subsection{Articles and their Stylistic Features}

\subsection{Articles and their Stylistic Features}
\label{sec:language}

The style in which news is presented to the reader
plays a pivotal role in understanding its credibility. The
desired property for news is to be objective and unbiased. In this section, we examine the different stylistic indicators of news credibility. All the lexicons used in this section are compiled from~\cite{recasens2013,mukherjee2014}.

%%%GW: drastically trimmed this !!!

\noindent{\bf Assertives}: 
Assertive verbs (e.g., ``claim'') complement and modify a proposition in a sentence. 
They capture the degree of certainty to which a proposition holds. % \cite{recasens2013}. 

\noindent{\bf Factives}: Factive verbs (e.g., ``indicate'')
pre-suppose the truth of a proposition in a sentence.
%%%GW: "presuppose" is not clear to me - can we say this differently?
%%%subho: take for granted, presume

\noindent{\bf Hedges}: These are mitigating words (e.g., ``may'') to soften the degree of
commitment to a proposition.% \cite{recasens2013}.
%Note the contrast between the situations presented by \emph{will} and \emph{may} in the following examples.
%%%GW: why do we say assertion here and proposition previously? any diff?

\noindent{\bf Implicatives}:
These words trigger pre-supposition in an utterance. For
example, usage of the word \emph{complicit} indicates participation in an activity in an unlawful way.

\noindent{\bf Report verbs}: These verbs (e.g., ``argue'')
are used to indicate the attitude towards the
source, or report what someone said more accurately,
rather than using just \emph{say} and \emph{tell}. 

\noindent{\bf Discourse markers}: These capture the degree of confidence, perspective, and certainty in the set of propositions made. For instance, strong modals (e.g., ``could''), 
probabilistic adverbs (e.g., ``maybe''), and conditionals (e.g., ``if'') depict a high degree of uncertainty and hypothetical situations, whereas weak modals (e.g., ``should'')
 and inferential conjunctions (e.g., ``therefore'') depict certainty.

\noindent{\bf Subjectivity and bias}:
News is supposed to be objective: writers should not
convey their own opinions, feelings or prejudices in their
stories. For example, a news titled ``Why do conservatives hate your children?'' is not considered %credible or
objective journalism. We use a subjectivity lexicon\footnote{\small \href{http://mpqa.cs.pitt.edu/lexicons/subj\_lexicon/}{http://mpqa.cs.pitt.edu/lexicons/subj\_lexicon/}}, a list of positive and negative
opinionated words\footnote{\small \href{http://www.cs.uic.edu/~liub/FBS/opinion-lexicon-English.rar}{http://www.cs.uic.edu/~liub/FBS/opinion-lexicon-English.rar}}, and an affective lexicon\footnote{\small \href{http://wndomains.fbk.eu/wnaffect.html}{http://wndomains.fbk.eu/wnaffect.html}} to detect subjective clues in articles.
The {\em affective features} capture the state of mind (like attitude and emotions) of the writer while writing an article or post (e.g., anxiousness, confidence, depression, favor, malice, sympathy etc.).

%{\tt GW: which lexcion(s)????? be explict!!!!!}

We additionally harness a lexicon of bias-inducing words
extracted from the Wikipedia edit history from~\cite{recasens2013} exploiting its Neutral Point of View Policy to keep its articles ``fairly, proportionately, and as far as possible without bias, all significant views that have been published by reliable sources on a {\em topic}''.

%{Wikipedia has a Neutral Point of View policy to keep its articles ``fairly, proportionately, and as far as possible without bias, all significant views that have been published by reliable sources on a topic". Accordingly, editors identify and rewrite biased passages that do not conform to this, and mark them with NPOV tags.} {\tt Subho: this can be left out in need for space}

%Table~\ref{tab:linguistic} shows an excerpt of the words belonging to
%each of the linguistic categories and the number of
%elements in each category.

%%%GW: the table adds little value to the text, but has the risk of leading the reader=reviewer on a secondary track
%%%GW: the count column is even "more secondary" as it simply gives dictionary sizes
%%%subho: should we give a snapshot of the table just to give a feel of the features and words used. In the experiments we show language features to work well, or we may give the snapshot during discussions?

\comm{
\begin{table}
\scriptsize
 \begin{tabular}{p{1.8cm}p{5cm}p{0.5cm}}
 \toprule
\textbf{Category} & \textbf{Elements} & \textbf{Count}\\\midrule
{Assertives} & think, believe, suppose, expect, imagine& 66\\
{Factives} & know, realize, regret, forget, find out & 27\\
{Hedges} &  postulates, felt, likely, mainly, guess& 100\\
{Implicatives} & manage, remember, bother, get, dare& 32\\
{Report} & claim, underscore, alert, express, expect & 181\\\midrule
{Subjectivity}&&\\
\myindent Bias & apologetic, summer, advance, cornerstone, & 354\\
\myindent Negative & hypocricy, swindle, unacceptable, worse & 4783\\
\myindent Positive & steadiest, enjoyed, prominence, lucky & 2006\\
\myindent Subj. Clues & better, heckle, grisly, defeat, peevish & 8221\\
\myindent Affective & disgust, gross\_out, revolt, repe & 2978\\
\midrule
Discourse &&\\
\begin{tabular}{ll}
\myindent Strong modals (might, could) & Weak modals (should, ought)\\
\myindent Conditionals (if) & Negation (no, not)\\
\myindent Inferential conj. (therefore, thus) & Contrasting conj. (despite, though)\\
\myindent Following conj. (but, however) & Definite det. (this, that)\\
\myindent First person (I, we) & Second person (you, your)\\
\myindent Third person (he, she) & Question particles (why, what)\\
\myindent Adjectives (correct, extreme) & Adverbs (maybe, about)\\
\bottomrule
\end{tabular}
\end{tabular}
\vspace{-0.5em}
\caption{Stylistic features for news credibility.}
\label{tab:linguistic}
\vspace{-1em}
\end{table}
}%\comment

\noindent {\bf Feature vector construction}: For each stylistic feature type $f_i$ 
%shown in Table \ref{tab:linguistic} 
and each news article $d_j$, we compute the relative frequency of words of type $f_i$ occurring in $d_j$, thus constructing a feature vector $F^L(d_j) =\langle freq_{ij} = \#(words~in~f_i) ~ / ~length(d_j) \rangle$. Consider the review $r_{j,k}$ written by user $u_k$ on the article $d_j$. For each such review, analogous to the per-article stylistic feature vector $\langle F^L(d_j) \rangle$, we construct a {\em per-review} feature vector $\langle F^L(r_{j,k}) \rangle$.

%\label{sec:language}

%\subsection{Articles and their Topics}
\subsection{Articles and their Topics}
\label{sec:topics}

\comm{
According to the Prominence-Interpretation theory~\cite{fogg2003} on how people assess credibility, there are several factors that affect prominence 
%(noticeability) 
of news: involvement of the user, content of the
news, task and experience of the user, and individual differences. The underlying \textit{topic} of news, and correspondingly the viewpoint of users influences all of these factors in different ways.
}%\comment

%An earlier work in content driven trust propagation~\cite{vg2011} found that news coverage of topics such as ``Bush administration" and ``WikiLeaks" are fairly trustworthy; whereas the coverage on ``Republican policy" and ``Democratic policy" are not considered as trustworthy by users. They also found the articles on ``Obama administration" to be significantly less trustworthy than the coverage of ``Bush administration".

%We observe that significant difference in opinion between different users on the credibility
%rating of articles arises due to conflict between their political viewpoint and that of the %article. We perform some preliminary experiments to establish this hypothesis, which we %discuss in Section~\ref{sec:hypotheses}.

Topic tags for news articles play an important role in user-perceived prominence, bias
and credibility, in accordance to the Prominence-Interpretation theory~\cite{fogg2003}. For example, the tag {\em Politics} is often viewed as an indicator
of potential bias and individual differences; whereas tags like {\em Energy} or {\em Environment} are perceived
as more neutral news and therefore invoke higher agreement in the community on the associated articles' credibility. Obviously, this can be misleading as there is a significant influence of Politics on all topics in all format of news.%irrespective of genre.
Certain
users have topic-specific expertise that make them rate
articles on those topics better than others. News
sources also have expertise on specific topics and provide
a better coverage of news on those topics than others. For example, National Geographic provides a good coverage
of news related to \textit{environment}, whereas The Wall Street
Journal provides a good coverage on \textit{economic} policies.

However, 
%$33.12\%$ of the 
%%% such specific percentages need a reference, but can still not be quantitatively interpreted here, hence commented out
many
news articles do not have any explicit topic
tag. In order to automatically identify the underlying theme of the article, we use Latent
Dirichlet Allocation (LDA)~\cite{blei2003} to learn the latent topic distribution in the corpus. LDA assumes a document to have a distribution over a set of topics, and each topic to have a distribution over words. 
Table~\ref{tab:lattopics} shows an excerpt of the top topic words in each topic,
where we manually added illustrative labels for the topics. The latent topics also capture some subtle themes not detected by the explicit tags. For example, \emph{Amy Goodman} is an American broadcast journalist, syndicated columnist and investigative reporter who is considered highly credible in the community. Also, associated with that topic cluster is \emph{Amanda Blackhorse}, a Navajo
activist and plaintiff in the Washington Redskins case.

\noindent{\bf Feature vector construction}: For each document $d_j$ and each of its review $r_{j,k}$, we create feature vectors $\langle F^T(d_j) \rangle$ and $\langle F^T(r_{j,k}) \rangle$ respectively, using the learned {\em latent} topic distributions, as well as the {\em explicit} topic tags. Section~\ref{subsec:latentTopics} discusses our method to learn the topic distributions.

%Another example latent topic shown there is \emph{Alternet} which is a progressive/liberal activist news service whose mission is to ``inspire citizen action and advocacy on the environment, human rights and civil liberties, social justice, media, and health care issues".

\begin{table}
\scriptsize
\centering
	\begin{tabular}{p{1.45cm}p{6.7cm}}
		\toprule
		\textbf{Latent Topics} & \textbf{Topic Words}\\\midrule
		Obama admin. & obama, republican, party, election, president, senate, gop, vote\\
		Citizen journ. & cjr, jouralism, writers, cjrs, marx, hutchins, reporting, liberty, guides\\
		US military & iraq, war, military, iran, china, nuclear, obama, russia, weapons\\
		AmyGoodman & democracy, military, civil, activist, protests, killing, navajo, amanda\\
		Alternet & media, politics, world news, activism, world, civil, visions, economy\\
		Climate & energy, climate, power, water, change, global, nuclear, fuel, warming\\
		\bottomrule
	\end{tabular}
	\vspace{-0.5em}
	\caption{Latent topics (with illustrative labels) and their words.}
	\label{tab:lattopics}
	\vspace{-2em}
\end{table}
%{\tt GW: trim the table: one row per topic!!!!!!}

%\label{sec:topics}

\subsection{News Sources}

A news source is considered \emph{trustworthy} if it generates highly \emph{credible} articles. We examine the effect of different features of a news source on its trustworthiness based on user assigned ratings in the community. 
%%%GW: the following repeats the intro
\comm{
Previous works~\cite{flanagin2000, johnson2007} 
found that people consider newspaper information to be more credible than online information, TV or magazines. However,  politically-interested internet users find online information to be more credible than traditional media counterparts. 
Besides the medium (print vs. broadcast vs. online), 
the level of experience of the users, and the format of news or type of information need plays a role in their perceived credibility of the news content. 
}%\comment
We consider the following source 
features (summarized in Table~\ref{tab:sources}):
the type of \emph{media} (e.g., online, newspaper, tv, blog), \emph{format} of news (e.g., news analysis, opinion, special report, news report, investigative report), (political) \emph{viewpoint} (e.g., left, center, right),
\emph{scope} (e.g., international, national, local), the top \emph{topics} covered by the source, and their topic-specific \emph{expertise}.
%The observations from our study are presented in %Section~\ref{subsec:qualanalysis}.

%%%GW: counts are unnecessary and potentially distracting
\begin{table}
\scriptsize
\centering
 \begin{tabular}{p{1.3cm}p{5.5cm}}
 \toprule
\textbf{Category} & \textbf{Elements} %& \textbf{Count}
\\\midrule
{Media} & newspaper, blog, radio, magazine, online %& 9
\\
{Format} & editorial, investigative report, news, research %& %26
\\
{Scope} & local, state, regional, national, international %& %6
\\
{Viewpoint} & far left, left, center, right, neutral %& %6
\\
{Top Topics} & politics, weather, war, science,, U.S. military %& %282
\\
{Expertise on Topics}& U.S. congress, Middle East, crime, presidential election, Bush administration, global warming %& %28
\\
\bottomrule
\end{tabular}
\vspace{-1em}
\caption{Features for source trustworthiness.}
\label{tab:sources}
\vspace{-2.5em}
\end{table}

%creating feature vectors for regression
\noindent{\bf Feature vector construction}: For each news source $s_l$, we create a feature vector $\langle F^S(s_l) \rangle$ using features in Table~\ref{tab:sources}. Each element $f_i^S(s_l)$ is $1$ or $0$ indicating presence or absence of a feature. Note that above features include the top (explicit) topics covered by any source, and its topic-specific expertise for a subset of those topics.
\label{sec:sources}

\vspace{-0.5em}
\subsection{Users, Ratings and Interactions}

A user's expertise in judging news credibility depends on many factors. \cite{einhorn1977} discusses the following traits for recognizing an expert.

%\squishlisttwo
% \item An expert user needs to be recognized by other members %in the community. This can be captured by how the community %members perceive the quality of the user's article ratings.
%  \item Experience is an uncertain indicator of user expertise. %\textit{Community engagement} of the member is taken as an %indicator of his experience.
% \item Intra-expert agreement should be high and different %experts should converge on the same signs.
% \item Experts should be independent of bias.
%\squishend
%%%GW: this itemized list is merely the preview of the following paragraphs, hence commented out

%We validate all these hypothesis in Section~\ref{sec:hypotheses}. %Or proposed model captures the following aspects of user %expertise in identifying credible news.
 
\noindent{\bf Community Engagement} of the user is an obvious measure for judging the user authority in the community. We capture this with different features: number of answers, 
ratings given, comments, ratings received, disagreement and number of raters.

%\noindent{\bf Intra-User Agreement}: 
\noindent{\bf Inter-User Agreement}: 
Expert users typically agree on what constitutes a credible article. This is inherently captured in the proposed graphical model, where a user gains expertise by assigning credibility ratings to articles that corroborate with other expert users.

\noindent{\bf Topical Perspective and Expertise}: The potential for harvesting user preference and expertise in topics for rating prediction of reviews has been demonstrated in~\cite{mukherjee2014JAST,mcauley2013}. For credibility analysis the model needs to capture the user's {\em perspective} and {\em bias} towards certain topics based on their political inclination that bias their ratings, and their topic-specific {\em expertise} that allows them to evaluate articles on certain topics better as ``Subject Matter Experts''. These are captured as {\em per-user} feature weights for the stylistic indicators and topic words in the language of user-contributed reviews.

%In this work, we do not distinguish between the fine-grained %aspects of topic influence on users. Instead, we capture them %together as {\em per topic} feature weights for {\em each} user %for explicit and latent topic representations.
%%%GW: this is very vague: what features do we introduce here ?????

\noindent{\bf Interactions}:\label{sec:interactions}
%There are different kinds of interactions at play between users and other facets in the community that can be harnessed to identify expert users and credible articles.
In a community, users can upvote ({\em digg, like, rate}) the ratings of users that they appreciate, and downvote the ones they do not agree with. High review ratings from expert users increase the value of a user; whereas low ratings bring down her expertise. Similar to this {\em user-user} interaction, there can be {\em user-article}, {\em user-source} and {\em source-article} interactions which are captured as edges in our graphical model (by construction).
Consider the following anecdotal example in the community showing an expert in nuclear energy \textit{downvoting} another user's rating on nuclear radiation:
{\small \em
``Non-expert: Interesting opinion about health risks of nuclear radiation, from a physicist at Oxford University. He makes some reasonable points ...\\
Low rating by expert to above review: Is it fair to assume that you have no background in biology or anything medical? While this story is definitely very important, it contains enough inaccurate and/or misleading statements...''
}

\noindent{\bf Feature vector construction}: For each user $u_k$, we create an engagement feature vector $\langle F^E(u_k) \rangle$. In order to capture user {\em subjectivity}, in terms of different stylistic indicators of credibility, we consider the {\em per-review} language feature vector $\langle F^L(r_{j,k}) \rangle$ of user $u_k$ (refer to Section~\ref{sec:language}). To capture {\em user perspective and expertise} on different topics, we consider the {\em per-review} topic feature vector $\langle F^T(r_{j,k}) \rangle$ of each user $u_k$.
% (refer to Section~\ref{sec:topics}).

\label{sec:users}
\\

%\subsection{Interaction: Sources, Articles and Users}
%\input{interaction}
\vspace{-0.5em}

\section{Joint Probabilistic Inference}
\label{sec:inference}

In this section we incorporate the discussed features and insights into a joint
probabilistic graphical model. The task is to identify credible news articles,
trustworthy news sources, and expert users {\em jointly} in a news community. Table~\ref{tab:symbol} summarizes the important notations used in this section.

%{\tt GW: this section is very heavy in notation. It would be great to add a reference table with all notation: Summary of variables and notation !!!!!}\\
\vspace{-0.5em}
\begin{table}
\scriptsize
\centering
\begin{tabular}{p{1.6cm}p{1.7cm}p{4.2cm}}
\toprule
\textbf{Variables} & \textbf{Type} & \textbf{Description}\\\midrule
$d_j$ & Vector & Document with sequence of words $\langle w \rangle$\\
$s$ & Vector & Sources\\
$u$ & Vector & Users\\
$r_{j,k}$ & Vector & Review by user $u_k$ on document $d_j$\\
& & \myindent with sequence of words $\langle w \rangle$\\
$y_{j,k}$ & Real Number & Rating of $r_{j,k}$\\
$z$ & Vector & Sequence of topic assignments for $\langle w \rangle$\\
$\text{SVR}_{u_k}, \text{SVR}_{s_i}$ & Real Number & SVR prediction for users,
sources,\\
$\text{SVR}_L, \text{SVR}_T$ & \myindent $\in$ [1 \ldots 5] & \myindent
language, and topics\\
$\Psi = f(\langle \psi_j \rangle) $ & Real Number & Clique potential with
$\psi_j = \langle y_j, s_i, d_j,$ $ \langle u_k \rangle, \langle r_{j,k} \rangle \rangle
 $ for clique of $d_j$\\
\myindent ${\lambda=}$ $\langle \alpha_u, \beta_s, \gamma_1, \gamma_2 \rangle$ &
Vector & Combination weights for users $\langle u \rangle$, sources $\langle s
\rangle$, language and topic models\\
$y_{n \times 1}$ & Vector & Credibility rating of documents $\langle d \rangle$\\
$X_{n \times m}$ & Matrix & Feature matrix with $m = |U|+|S| + 2$\\
$Q_{n \times n}$ & Diagonal Matrix & $f(\lambda)$\\
$b_{n \times 1}$ & Vector & $f(\lambda, X)$\\
$\Sigma_{n \times n}$ & {\scriptsize CovarianceMatrix} & $f(\lambda)$\\
% & Matrix & \\
$\mu_{n \times 1}$ & Mean Vector & $f(\lambda, X)$\\
\bottomrule
\end{tabular}
\vspace{-1em}
\caption{Symbol table.}
\label{tab:symbol}
\vspace{-2em}
\end{table}

\subsection{Topic Model}
\label{subsec:latentTopics}

Consider an article $d$ consisting of a sequence of $\{N_d\}$ words denoted by
${w_1,w_2,...w_{N_d}}$. Each word is drawn from a vocabulary $V$ having unique
words indexed by ${1,2,...V }$. Consider a 
%sequence 
set of topic assignments $z = \{z_1,z_2,...z_K\}$ 
for $d$, 
%%%GW: these are the topics for d - right?
%%%this is what the wording "topic assignment" suggests to me
%%%otherwise I would talk about "set of topics"
where
each topic $z_i$ can be from a set of $K$ possible topics. 

LDA~\cite{blei2003} assumes each document $d$ to be associated with a multinomial distribution $\theta_d$ over topics $Z$ with a symmetric dirichlet prior $\rho$. $\theta_d(z)$ denotes the probability of occurrence of topic $z$ in document $d$. Topics have a multinomial distribution $\phi_z$ over words drawn from a vocabulary $V$ with a symmetric dirichlet prior $\zeta$. $\phi_z(w)$ denotes the probability of the word $w$ belonging to the topic $z$.
Exact inference is not possible due to intractable coupling between $\Theta$ and $\Phi$. We use Gibbs sampling for approximate inference.

Let $n(d, z, w)$ denote the count of the word $w$ occurring in document $d$ belonging to the topic $z$. In the following equation, $(.)$ at any position in the above count indicates marginalization, i.e.,  summing up the counts over all values for the corresponding position in $n(d,z,w)$.
The conditional distribution for 
%the update of 
the latent variable $z$ (with components $z_1$ to $z_K$)
is given by:

\vspace{-1em}
\begin{equation}
\label{eq.3}
\small
\begin{aligned}
 P(z_i=k| &w_i=w, z_{-i}, w_{-i}) \propto\\
  \frac{n(d, k, .) + \rho}{\sum_{k}n(d, k, .) + K \rho} &\times \frac{n(., k, w) + \zeta}{\sum_{w}n(., k, w) + V \zeta}
 \end{aligned}
\end{equation}
\vspace{-0.5em}
%%%GW: I moved propto to the first line, as it is easy to confuse propto with some coefficient rho and then believing that an equality sign is missing

Let $\langle T^E \rangle$ and $\langle T^L \rangle$ be the set of explicit topic tags and latent topic dimensions, respectively. The topic feature vector $\langle F^T \rangle$ for an article or review
combines both explicit tags and latent topics and is constructed as follows:

\vspace{-1.5em}
\begin{equation*}
\label{eq:topic}
\small
F^T_t(d) =
\begin{cases}
\#freq(w,d),& \text{if} \ T^E_{t^{'}}=F^T_t \\
\#freq(w,d) \times \phi_{T^L_{{t'}}}(w),& \text{if} \ T^L_{t^{'}} = F^T_t \text{ and } \phi_{T^L_{t^{'}}}(w) > \delta\\
0 & \text{otherwise}
\end{cases}\hfill
\end{equation*}
\vspace{-1em}

So for any word in the document matching an explicit topic tag, the corresponding element in the feature vector $\langle F^T \rangle$ 
is set to its occurrence count in the document. If the word belongs to any latent topic with probability greater than threshold $\delta$, the probability of the word belonging to that topic ($\phi_{t}(w)$) is added to the corresponding element in the feature vector, and set to $0$ otherwise.

\subsection{Support Vector Regression}
\label{subsec:SVR}

We use Support Vector Regression (SVR)~\cite{drucker1996} to combine the different features discussed in Section~\ref{sec:newscom}. SVR is an extension of the max-margin framework for SVM classification to the regression problem. It solves the following optimization problem to learn weights $w$ for features $F$:

{
\vspace{-2em}
\small
\begin{multline}
\label{eq.5}
 \min_{w} \frac{1}{2}  {w} ^T{w} + C \times
 \sum_{d=1}^{N} (max(0, |y_d - w^TF| - \epsilon))^2 \hfill
%   \text{where $\langle .,. \rangle$ denotes a scalar product.}\hfill
 \end{multline}
 \vspace{-1em}
}

%{\tt GW: why the special notation for scalar product -- w**T*w is also a scalar product ????? we should not use two different notations !!!!!}

\noindent \textbf{Article Stylistic Model}: We learn a stylistic regression model $\text{SVR}_L$ using the {\em per-article} stylistic feature vector $\langle F^L(d_j) \rangle$ for article $d_j$ (or, $\langle F^L(r_{j,k}) \rangle$ for review $r_{j,k}$), with the overall article rating $y_j$ (or, $y_{j,k}$) as the response variable.

\noindent \textbf{Article Topic Model}: Similarly, we learn a topic regression model $\text{SVR}_T$ using the {\em per-article} topic feature vector $\langle F^T(d_j) \rangle$ for article $d_j$ (or, $\langle F^T(r_{j,k}) \rangle$ for review $r_{j,k}$), with the overall article rating $y_j$ (or, $y_{j,k}$) as the response variable.

\noindent \textbf{Source Model}: We learn a source regression model $\text{SVR}_{s_i}$ using the {\em per-source} feature vector $\langle F^S(s_i) \rangle$ for source $s_i$, with the overall source rating as the response variable .

\noindent \textbf{User Model}: For each user $u_k$, we learn a user regression model $\text{SVR}_{u_k}$ with her {\em per-review} stylistic and topic feature vectors \\$\langle F^L(r_{j,k}) \cup F^T(r_{j,k}) \rangle$ for review $r_{j,k}$ for article $d_j$, with her overall review rating $y_{j,k}$ as the response variable.

Note that we use overall article rating to train article stylistic and topic models. For the user model, however, we take user assigned article ratings and per-user features. This model captures user subjectivity and topic perspective. The source models are trained on news-source specific meta-data and its ground-truth ratings.

\subsection{Continuous Conditional Random Field}

%As outlined in Section~\ref{sec:newscom}, 
We model our learning task as a Conditional Random Field (CRF), where the random variables are the ratings of news articles $\langle d_j \rangle$, news sources $\langle s_i \rangle$, users $\langle u_k \rangle$, and reviews $\langle r_{j,k} \rangle$. The objective is to predict the credibility ratings $\langle y_j \rangle$ of the articles $\langle d_j \rangle$.

%Let $\langle u^j_k \rangle$ denote the set of users reviewing article $d_j$.

The cliques in the CRF consist of an article $d_j$, its source $s_i$, set of users $\langle u_k \rangle$ reviewing it, and the corresponding user reviews $\langle r_{j,k} \rangle$ --- where $r_{j,k}$ denotes the review by user $u_k$ on article $d_j$. Different cliques are connected via the common news sources, and users. There are as many cliques as the number of news articles.

Let $\psi_j(y_j, s_i, d_j, \langle u_k \rangle, \langle r_{j,k} \rangle)$ be a
potential function for clique $j$. Each clique has a set of associated {\em
vertex} feature functions.
%$F_j$ with a weight vector $W$.
In our problem setting, we associate features to each vertex.% and ignore the edge feature functions.
%We denote the individual features and their weights as $f_{m,n}$ and $w_{n}$.
The features constituted by the stylistic, topic, source and user features explained in Section~\ref{sec:newscom} are:
%
%\begin{equation} \label{eq3}
$ F^L(d_j) \cup F^T(d_j) \cup F^S(s_i) \cup_k (F^E(u_k) \cup F^L(r_{j,k}) \cup F^T(r_{j,k})).$
%\end{equation}

%In the following, $y, \mu, b$ are vectors, and $X, Q, \Sigma$ are matrices.

A traditional CRF model allows us to have a {\em binary} decision if a news article is {\em credible} ($y_j=1$) or not ($y_j=0$), by estimating the conditional distribution with the probability {\em mass} function of the discrete random variable $y$:

\vspace{-1.5em}
\begin{equation} \label{eq4}
\small
 Pr( y | D, S, U, R) = \frac{\prod_{j=1}^{n} exp(\psi_j(y_j, s_i, d_j, \langle u_k 
\rangle, \langle r_{j,k} \rangle))}{{\sum_{y} \prod_{j=1}^{n} exp(\psi_j(y_j, s_i,
d_j,                                                                        
\langle u_k \rangle, \langle r_{j,k} \rangle))}}
\end{equation}
\vspace{-1.5em}

But in our problem setting, we want to estimate the credibility {\em rating} of an article. Therefore, we need to estimate the conditional distribution with the probability {\em density} function of the continuous random variable $y$:\\
%{\tt GW: this sounds odd and confusing -- the equation for the continuous case is still a conditional distribution, not the density function ?????}\\
%{\tt GW: why are the bounds of the integral non-standard: from plus infinity to minus infinity ?????}\\

{
\vspace{-3em}
\small
\begin{equation}
\setlength{\mathindent}{0pt}
\label{eq.5}
 Pr( y | D, S, U, R) = \frac{\prod_{j=1}^{n} exp(\psi_j(y_j, s_i, d_j, \langle u_k
\rangle, \langle r_{j,k} \rangle))}{\int_{-\infty}^{\infty} \prod_{j=1}^{n}
exp(\psi_j(y_j, s_i, d_j, \langle u_k \rangle, \langle r_{j,k} \rangle))dy}\\
\end{equation}
\vspace{-2em}
}

\begin{comment}
where the potential function made explicit in terms of features and weights as:

\begin{equation}
\small
\label{eq.6}
\begin{aligned}
\psi_j(s_i, d_j, \langle u_k \rangle, \langle r_{j,k} \rangle) &= exp (\sum_n
w_n \times f_{m,n}(s_i, d_j, \langle u_k \rangle, \langle r_{j,k} \rangle))
 \end{aligned}
\end{equation}
\end{comment}

Given a news article $d_j$, its source id $s_i$, and a set of user ids $\langle u_k \rangle$ who reviewed the article, the regression models $\text{SVR}_L(d_j)$, $\text{SVR}_T(d_j)$, $\text{SVR}_{s_i}$, $\langle \text{SVR}_{u_k}(d_j) \rangle$ (discussed in Section~\ref{subsec:SVR}) predict rating of $d_j$. For notational brevity, hereafter, we drop the argument $d_j$ from the SVR function.
%%%GW: avoid repeating the features all the time
%based on its stylistic and topic features extracted from text, %source features extracted from the meta-data of its source, and %user models learned from the training data. 
These SVR predictors are for separate feature groups and
independent of each other. Now we combine the 
different SVR models to capture mutual interactions, such that the weight for each SVR model reflects our confidence on its quality. 
Errors by an SVR are penalized by the squared loss between the predicted article rating and the ground-truth rating. There is an additional constraint that for any clique {\em only} the regression models corresponding to the news-source and users present in it should be activated. This can be thought of as partitioning the input feature space into subsets, with the 
features inside a clique capturing {\em local} interactions, and the {\em global} weights capture the overall quality of the random variables via the shared information between the cliques (in terms of common sources, users, topics and language features) --- an ideal setting for using a CRF. Equation~\ref{eq.7} shows one such linear combination. 
Energy function of an individual clique is given by:

{
\vspace{-1.6em}
\setlength{\belowdisplayskip}{0pt} \setlength{\belowdisplayshortskip}{0pt}
\setlength{\abovedisplayskip}{0pt} \setlength{\abovedisplayshortskip}{0pt}
\setlength{\mathindent}{0pt}
\small
\centering
\begin{multline}
\label{eq.7}
\psi_j(y_j, s_i, d_j, \langle u_k \rangle, \langle r_{j,k} \rangle) =
-\sum_{k=1}^{k=U} \alpha_{k} \mathbb{I}_{u_k}(d_j) (y_j - \text{SVR}_{u_k})^2 \\ -
\sum_{i=1}^{i=S} \beta_i \mathbb{I}_{s_i}(d_j) (y_j - \text{SVR}_{s_i})^2
 -\gamma_1 (y_j - \text{SVR}_L)^2 -\gamma_2 (y_j - \text{SVR}_T)^2
\end{multline}
}
\vspace{-2.5em}
 
{\noindent Indicator functions $\mathbb{I}_{u_k}(d_j)$ and $\mathbb{I}_{s_i}(d_j)$ are 1 if $u_k$ is a reviewer and $s_i$ is the source of article $d_j$ respectively, and are $0$ otherwise.}

As the output of the SVR is used as an input to the CCRF in Equation~\ref{eq.7}, each element of the input feature vector is already predicting the output variable.  The learned parameters $\lambda = \langle \alpha, \beta, \gamma_1, \gamma_2 \rangle$ (with dimension$(\lambda)=|U|+|S|+2$) of the linear combination of the above features depict how much to trust individual predictors. Large $\lambda_k$ on a particular predictor places large penalty on the mistakes committed by it, and therefore depicts a higher quality for that predictor. $\alpha_u$ corresponding to user $u$ can be taken as a proxy for that user's \textit{expertise}, allowing us to obtain a ranked list of expert users.
Similarly, $\beta_s$ corresponding to news source $s$ can be taken as a proxy for that source's \textit{trustworthiness}, allowing us to obtain a ranked list of trustworthy news sources.

Overall energy function of all cliques is given by:
%{\tt GW: from here on it gets a bit messy -- regarding layout of equations and, especially, lack of guiding the reader through the math!!!!!}

{
\vspace{-1em}
\small
\centering
\setlength{\mathindent}{0pt}
\setlength{\belowdisplayskip}{0pt} \setlength{\belowdisplayshortskip}{0pt}
\setlength{\abovedisplayskip}{0pt} \setlength{\abovedisplayshortskip}{0pt}
\begin{multline*}
\myindent \myindent \myindent \myindent
 \Psi = \sum_{j=1}^{n} \psi_j(y_j, s_i, d_j, \langle u_k \rangle, \langle r_{j,k}
\rangle)\\
\vspace{-1em}
 \text{\normalsize (Substituting $\psi_j$ from Equation~\ref{eq.7} and
re-organizing terms)}\\
     \Psi = \sum_{j=1}^{n} (-\sum_{k=1}^{k=U} \alpha_{k} \mathbb{I}_{u_k}(d_j)
(y_j - \text{SVR}_{u_k})^2 \\ - \sum_{i=1}^{i=S} \beta_i \mathbb{I}_{s_i}(d_j)
(y_j - \text{SVR}_{s_i})^2 -\gamma_1 (y_j - \text{SVR}_L)^2 -\gamma_2 (y_j - \text{SVR}_T)^2)\\
      = -\sum_{j=1}^{n} y_j^2 [\sum_{k=1}^{k=U} \alpha_{k} \mathbb{I}_{u_k}(d_j)
+ \sum_{i=1}^{i=S} \beta_i \mathbb{I}_{s_i}(d_j) + \gamma_1 + \gamma_2] \\
      + \sum_{j=1}^{n} 2y_j[\sum_{k=1}^{k=U} \alpha_{k} \mathbb{I}_{u_k}(d_j)
\text{SVR}_{u_k} + \sum_{i=1}^{i=S} \beta_i \mathbb{I}_{s_i}(d_j)\text{SVR}_{s_i} + \gamma_1
\text{SVR}_L \\
      + \gamma_2 \text{SVR}_T] -\sum_{j=1}^{n} [\sum_{k=1}^{k=U} \alpha_{k}
\mathbb{I}_{u_k}(d_j) \text{SVR}_{u_k}^2 + \sum_{i=1}^{i=S} \beta_i
\mathbb{I}_{s_i}(d_j)\text{SVR}_{s_i}^2 \\+ \gamma_1 \text{SVR}_L^2
      + \gamma_2 \text{SVR}_T^2]
\end{multline*}
\begin{multline*}
\text{\normalsize Organizing the bracketed terms into variables as follows:}\\
Q_{i,j} =
\begin{cases}
\sum_{k=1}^{k=U} \alpha_{k}\mathbb{I}_{u_k}(d_i) + \sum_{l=1}^{l=S} \beta_l \mathbb{I}_{s_l}(d_i) + \gamma_1 + \gamma_2 & i=j \\
0 & \ i \neq j
\end{cases}\\
b_i = 2 [\sum_{k=1}^{k=U} \alpha_{k} \mathbb{I}_{u_k}(d_i) \text{SVR}_{u_k} + \sum_{l=1}^{l=S} \beta_l \mathbb{I}_{s_l}(d_i) \text{SVR}_{s_l}+ \gamma_1 \text{SVR}_L + \gamma_2 \text{SVR}_T]\\
c = \sum_{j=1}^{n} [\sum_{k=1}^{k=U} \alpha_{k} \mathbb{I}_{u_k}(d_j)
\text{SVR}_{u_k}^2 + \sum_{i=1}^{i=S} \beta_i \mathbb{I}_{s_i}(d_j)\text{SVR}_{s_i}^2 +
\gamma_1 \text{SVR}_L^2 + \gamma_2 \text{SVR}_T^2]
\end{multline*}
}

We can derive:

{
\vspace{-1em}
\setlength{\belowdisplayskip}{0pt} \setlength{\belowdisplayshortskip}{0pt}
\setlength{\abovedisplayskip}{0pt} \setlength{\abovedisplayshortskip}{0pt}
\begin{multline}
 \centering
 \small
 \Psi =  -y^TQy +y^Tb - c \hfill
\end{multline}
}

\vspace{-0.5em}

\noindent Substituting $\Psi$ in Equation~\ref{eq.5}:

{
\setlength{\mathindent}{0pt}
\small
\centering
\vspace{-0.5em}
\setlength{\belowdisplayskip}{0pt} \setlength{\belowdisplayshortskip}{0pt}
\setlength{\abovedisplayskip}{0pt} \setlength{\abovedisplayshortskip}{0pt}
\begin{equation}
\label{eq.9}
\begin{split}
  P(y|X) &= \frac{\prod_{j=1}^{n} exp(\psi_j)}{\int_{-\infty}^{\infty}
\prod_{j=1}^{n} exp(\psi_j) dy}\\
 &= \frac{exp(\Psi)}{\int_{-\infty}^{\infty}  exp(\Psi) dy}\\
 &= \frac{exp(-y^TQy +y^Tb)}{\int_{-\infty}^{\infty} exp(-y^TQy +y^Tb) dy}\\
 &= \frac{exp(-\frac{1}{2}y^T\Sigma^{-1}y+y^T\Sigma^{-1}\mu)}{\int_{-\infty}^{\infty} exp(-\frac{1}{2}y^T\Sigma^{-1}y+y^T\Sigma^{-1}\mu)dy}
 \end{split}
\end{equation}
}

\myindent \myindent \myindent{\normalsize (Substituting $Q = \frac{1}{2}\Sigma^{-1}, b = \Sigma^{-1}\mu$)}

\vspace{0.5em}
%Making the substitutions:
%$Q = \frac{1}{2}\Sigma^{-1}, b = \Sigma^{-1}\mu,\\ \int_{-\infty}^{\infty}exp(-\frac{1}{2}y^T\Sigma^{-1}y+y^T\Sigma^{-1}\mu)dy = \frac{(2\pi)^{n/2}}{|\Sigma^{-1}|^{\frac{1}{2}}}exp(\frac{1}{2}\mu^T\Sigma^{-1}\mu)$, Eqn.~\ref{eq.9} can be transformed into a multivariate Gaussian distribution:
%{\tt GW: not clear what you are talking about here !!!!!}\\
\noindent Equation~\ref{eq.9} can be transformed into a multivariate Gaussian distribution after substituting $\int_{-\infty}^{\infty}exp(-\frac{1}{2}y^T\Sigma^{-1}y+y^T\Sigma^{-1}\mu)dy = \frac{(2\pi)^{n/2}}{|\Sigma^{-1}|^{\frac{1}{2}}}exp(\frac{1}{2}\mu^T\Sigma^{-1}\mu)$. Therefore obtaining,

{
\small
%\vspace{-1em}
\setlength{\belowdisplayskip}{0pt} \setlength{\belowdisplayshortskip}{0pt}
\setlength{\abovedisplayskip}{0pt} \setlength{\abovedisplayshortskip}{0pt}
 \begin{multline}
 \label{eq.8}
 P(y|X) = \frac{1}{{(2\pi)}^{\frac{n}{2}}{|\Sigma|}^\frac{1}{2}}exp(-\frac{1}{2}(y-\mu)^T\Sigma^{-1}(y-\mu)) \hfill
\end{multline}
}

$Q$ represents the contribution of $\lambda$ to the covariance matrix $\Sigma$. Each row of the vector $b$ and matrix $Q$ corresponds to one training instance, representing the {\em active} contribution of features present in it.
To ensure Equation~\ref{eq.8} represents a valid Gaussian distribution, the covariance matrix $\Sigma$ needs to be positive definite for its inverse to exist. For that the diagonal matrix $Q$ needs to be a positive semi-definite matrix. This can be ensured by making all the diagonal elements in $Q$ greater than $0$, by constraining $\lambda_k > 0$.

Since this is a constrained optimization problem, gradient ascent cannot be directly used. We follow the approach similar to~\cite{radosavljevicECAI2010} and maximize log-likelihood with respect to $log\ \lambda_k$, instead of $\lambda_k$ as in standard gradient ascent, making the optimization problem unconstrained as:

{
\vspace{-0.5em}
\small
\setlength{\mathindent}{0pt}
\setlength{\belowdisplayskip}{0pt} \setlength{\belowdisplayshortskip}{0pt}
\setlength{\abovedisplayskip}{0pt} \setlength{\abovedisplayshortskip}{0pt}
\begin{multline}
\label{eq.11}
 \frac{\partial logP(y|X)}{\partial log\lambda_k} = \alpha_k(\frac{\partial logP(y|X)}{\partial \lambda_k})\hfill
 \end{multline}

 \vspace{0.5em}
 \text{\normalsize Taking partial derivative of the $log$ of Equation~\ref{eq.8} w.r.t $\lambda_k$}:
 
 \begin{multline}
\frac{\partial logP(y|X)}{\partial \lambda_k} = \frac{1}{2} \frac{\partial}{\partial \lambda_k} (-y^T\Sigma^{-1}y +2y^T\Sigma^{-1}\mu - \mu^T \Sigma^{-1} \mu \\ + log|\Sigma^{-1}| + Constant)\hfill
\end{multline}
\vspace{-0.5em}
}

Substituting the following in the above equation:

{
%\vspace{-1em}
\small
\setlength{\belowdisplayskip}{0pt} \setlength{\belowdisplayshortskip}{0pt}
\setlength{\abovedisplayskip}{0pt} \setlength{\abovedisplayshortskip}{0pt}
\centering
\begin{equation*}
\small
\centering
 \begin{split}
  \frac{\partial \Sigma^{-1}}{\partial \lambda_k} &= 2\frac{\partial Q}{\partial \lambda_k}\\
&= 2I\\
\frac{\partial \Sigma^{-1}\mu}{\partial \lambda_k} &= \frac{\partial b}{\partial \lambda_k} \ \ [\because \mu = \Sigma b]\\
&= 2X_{(.),k}
\end{split}
\end{equation*}
{\normalsize where, $X_{(.),k}$ indicates the $k^{th}$ column of the
feature matrix $X$.}

\begin{equation*}
\begin{split}
\frac{\partial \Sigma}{\partial \lambda_k} &= -\Sigma \frac{\partial \Sigma^{-1}}{\partial \lambda_k}\Sigma \\
&= - 2\Sigma\Sigma\\
\frac{\partial}{\partial \lambda_k}(\mu^T \Sigma^{-1} \mu) &= \frac{\partial}{\partial \lambda_k}(b^T \Sigma b) \\
&= b^T\frac{\partial \Sigma b}{\partial \lambda_k} + \frac{\partial b^T}{\partial \lambda_k}\Sigma b\\
&= b^T(\Sigma \frac{\partial b}{\partial \lambda_k} + \frac{\partial \Sigma}{\partial \lambda_k}b) + \frac{\partial b^T}{\partial \lambda_k}\Sigma b\\
&= 4 X_{(.),k}\Sigma b -2b^T\Sigma \Sigma b \\
&= 4X_{(.),k} \mu - 2\mu^T\mu\\
\frac{\partial log|\Sigma^{-1}|}{\partial \lambda_k}&=\frac{1}{|\Sigma^{-1}|}\text{Trace}(|\Sigma^{-1}|\Sigma\frac{\partial
\Sigma^{-1}}{\partial \lambda_k})\\
&= 2\text{Trace}(\Sigma)\hfill
 \end{split}
\end{equation*}
}

We can derive the gradient vector:

{
\setlength{\belowdisplayskip}{0pt} \setlength{\belowdisplayshortskip}{0pt}
\setlength{\abovedisplayskip}{0pt} \setlength{\abovedisplayshortskip}{0pt}
\begin{equation}
\small
\frac{\partial logP(y|X)}{\partial \lambda_k} = -y^Ty + 2y^TX_{(.),k} -
2X^T_{(.),k}\mu + \mu^T\mu + \text{Trace}(\Sigma)
\end{equation}
\vspace{-0.5em}
}

Let $\eta$ denote the learning rate. The update equation is given by:

{
\setlength{\belowdisplayskip}{0pt} \setlength{\belowdisplayshortskip}{0pt}
\setlength{\abovedisplayskip}{0pt} \setlength{\abovedisplayshortskip}{0pt}
\begin{equation}
\centering
\small
log\lambda_k^{new} = log\lambda_k^{old} + \eta \frac{\partial logP(y|X)}{\partial log\lambda_k} 
\end{equation}
}

Once the model parameters are learned using gradient ascent, the inference for the prediction $y$ of the article credibility rating is straightforward. As we assume the distribution to be Gaussian, the prediction is the expected value of the function, given by the mean of the distribution:
$ y\prime = argmax_y\ P(y|X) = \mu = \Sigma  b$.\\
Note that $\Sigma$ and $b$ are both a function of $\lambda= \langle \alpha, \beta, \gamma_1, \gamma_2 \rangle$ which represents the combination weights of various factors to capture mutual interactions. The optimization problem determines the optimal $\lambda$ for reducing the error in prediction.
%{\tt GW: I am confused by this equation: if this is equal to Sigma*b, isn't this a constant solution given the input???? Sigma and b are input values, right???? even if I'm misinterpreting something here, this indicates that we need better reader guidance!!!!!}

\vspace{-0.5em}

%\section{Dataset Statistics}
\section{Use Case: Newstrust}

We performed experiments with data from a typical news community: {\em newstrust.net}\footnote{Code and data available at\\ {\href{http://www.mpi-inf.mpg.de/impact/credibilityanalysis/}{http://www.mpi-inf.mpg.de/impact/credibilityanalysis/}}}.
This community is similar to {\em digg.com} and {\em reddit.com}, but has more refined ratings and interactions. We chose NewsTrust 
because of the availability of {\em ground-truth} ratings for credibility analysis of news articles;
such ground-truth is not available for the other communities.

We collected {\em stories} from NewsTrust from May, 2006 to May, 2014. Each such story features a {\em news article} from a source (E.g. BBC, CNN, Wall Street Journal) that is posted by a member, and reviewed by other members, many of whom are {\em professional journalists} and {\em content experts}\footnote{http://www.newstrust.net/help\#about\_newstrust}. We crawled all the stories with their explicit topic tags and other associated meta-data.
We crawled all the {\em news articles} from their original sources that were featured in any NewsTrust story. The earliest story dates back to May 1, 1939 and the latest one is in May 9, 2014.

We collected all {\em member profiles} containing information about the demographics, occupation and expertise of the members along with their activity in the community in terms of the posts, reviews and ratings; as well as \textit{interaction} with other members. The members in the community can also rate each others' ratings. The earliest story rating by a member dates back to May, 2006 and the most recent one is in Feb, 2014.
In addition, we collected information on member evaluation of news sources, and other information (e.g., type of media, scope, viewpoint, topic specific expertise) about source from its {\em meta data}.

\noindent{\bf Crawled dataset:} Table~\ref{tab:data} shows the dataset statistics. In total $~62$K unique news articles were reviewed in NewsTrust in the given period, out of which we were able to extract $~47$K full articles from the original sources like New York Times, TruthDig, ScientificAmerican etc --- a total of $~5.6$K distinct sources. The remaining articles were not available for crawling. There are $~84.7$K stories featured in NewsTrust for all the above articles,
% (including a few re-submissions of news articles which are re-reviewed by different sections of the 
%community), 
out of which $~52.5K$ stories refer to the news articles we managed to extract from their original sources. The average number of reviews per story is $1.59$. For general analysis we use the entire dataset. For experimental evaluation of the CCRF and hypotheses testing, we use only those stories ($~18.5$K) with a {\em minimum of $3$ reviews} that refer to the news articles we were able to extract from original sources.

\begin{table}
\scriptsize
\centering
\begin{tabular}{lr}
\toprule
\textbf{Factors} & \textbf{Count}\\\midrule
Unique news articles reviewed in NewsTrust & 62,064\\
NewsTrust stories on news articles & 84,704\\
NewsTrust stories with $\geq 1$ reviews & 43,107\\
NewsTrust stories with $\geq 3$ reviews & 18,521\\
%28,211, 18,521\\
NewsTrust member reviews of news articles & 134,407\\
\midrule
News articles extracted from original sources & 47,565\\
NewsTrust stories on extracted news articles & 52,579\\
\midrule
News sources & 5,658\\
Journalists who wrote news articles & 19,236\\
Timestamps (month and year) of posted news articles & 3,122\\
\midrule
NewsTrust members who reviewed news articles & 7,114\\
NewsTrust members who posted news articles & 1,580\\
\midrule
News sources reviewed by NewsTrust members & 668\\
\midrule
Explicit topic tags & 456\\
Latent topics extracted & 300\\
\bottomrule
\end{tabular}
\vspace{-.5em}
\caption{Dataset statistics.}
\label{tab:data}
\vspace{-1em}
\end{table}

\noindent{\bf Generated graph:} Table~\ref{tab:graph} shows the statistics of the graph
%\footnote{Analysis done using {http://gephi.github.io/}} 
constructed by the method of Section~\ref{sec:newscom}. 
%The graph has high connectivity and overlapping modular network structure.

%{\tt GW: fix/trim table: keep only essential numbers, check weakly CC's (appears twice), fix decimal commas  (5,21,630  should probably be 521,630 - or is there a digit missing?}\\
\begin{table}
\scriptsize
\centering
\begin{tabular}{lrlr}
\toprule
\textbf{Factors} & \textbf{Count} & \textbf{Factors} & \textbf{Count}\\\midrule
Nodes & 181,364 & No. of weakly connected components & 12\\
\myindent Sources & 1,704 & Diameter & 8\\
\myindent Members & 6,906 &  Average path length & 47 \\
\myindent News articles & 42,204 & Average degree & 6.641\\
\myindent Reviews & 130,550 & Average clustering coefficient & 0.884\\
Edges & 602,239 & Modularity & 0.516\\
Total triangles & 521,630 & &\\
\bottomrule
\end{tabular}
\vspace{-1em}
\caption{Graph statistics.}
\label{tab:graph}
\vspace{-2em}
\end{table}

\noindent{\bf Ground-Truth for evaluation}: The members in the community can rate the credibility of a news article on a scale from $1$ to $5$ regarding $15$ qualitative aspects like facts, fairness, writing style and insight, and popularity aspects like recommendation, credibility and views.
Members give an overall \emph{recommendation} for the article explained to them as:
\textit{\small ``... Is this quality journalism? Would you recommend this story to a friend or colleague? ... This question is similar to the up and down arrows of popular social news sites like Digg and Reddit, but with a focus on quality journalism."}
Each article's aspect ratings by different members are weighted (and aggregated) by NewsTrust based on findings of \cite{lampe2007}, and the member expertise and member level (described below).
%Ratings weigh more if the reviewer is expert on a topic and receives positive ratings from other users.
This overall article rating is taken as the ground-truth for the article \emph{credibility} rating in our work.
A user's member level is calculated by NewsTrust
based on her community engagement, experience, other users' feedback on her ratings, profile transparency and validation by NewsTrust staff. This member level is taken as the proxy for user \emph{expertise} in our work.
Members rate news sources while reviewing an article. These ratings are aggregated for each source, and taken as a proxy for the source \emph{trustworthiness} in our work.

\noindent {\bf Training data:}
%{\tt GW: we need to explicitly say which parts of that data are used for training and which for testing. I suppose it's 90percent of everything for training, 10percent for testing, and then we do 10-fold cross-validation - correct? We can put this here, at the end of the Use Case section, or at the beginning of the experimental section, before 5.1 starts!!!!!}\\
We perform $10$-fold cross-validation {on the news articles}. During training on any $9$-folds of the data, the algorithm learns the user, source, language and topic models from user-assigned ratings to articles and sources present in the train split. We combine sources with less than $5$ articles and users with less than $5$ reviews into background models for sources and users, respectively.
This is to avoid modeling from sparse observations, and to reduce dimensionality of the feature space. However, while testing on the remaining {\em blind} $1$-fold we use {\em only the ids} of sources and users reviewing the article; we do not use any user-assigned ratings of sources or articles. For a new user and a new source, we draw parameters from the user or source background model. The results are averaged by $10$-fold cross-validation, and presented in the next section.

\noindent {\bf Experimental settings:} In the first two experiments we want to
find the power of the CCRF in predicting user rating behavior, and credibility
rating of articles. Therefore, the evaluation measure is taken as the
\textit{Mean Squared Error} (MSE) between the prediction and the actual
ground-rating in the community. For the latter experiments in finding expert
users (and, trustworthy sources) there is no absolute measure for predicting
user (and, source) quality; it only makes sense to find the relative ranking of
users (and, sources) in terms of their expertise (and, trustworthiness).
Therefore, the evaluation measure is taken as the {\em Normalized Discounted
Cumulative Gain} (NDCG) \cite{Jarvelin:TOIS2002} between the ranked list of users
(and, sources) obtained from CCRF and their actual ranking in the
community.\\
%%%GW: the hypotheses section is mostly emirical discussion of the data - hence dropped for CIKM
%\input{hypotheses}
\vspace{-0.5em}

\vspace{-1em}
%\section{Evaluation}
\section{Experiments}
\label{sec:evaluations}

\begin{comment}
\begin{table}[t]
\vspace{-0.5em}
\small
\centering
%\begin{minipage}{0.48\textwidth}
\begin{tabular}{p{3cm}p{1cm}}
\toprule
\textbf{Factors} & \textbf{\#Count}\\\midrule
News sources & 704\\
News articles & 11,432\\
Members & 5,177\\
Reviews & 61,499\\\bottomrule
\end{tabular}
\vspace{-0.5em}
\caption{Evaluation data statistics}
\label{tab:evaldata}
\end{comment}

\begin{table}
\vspace{-0.5em}
\small
\centering
\begin{tabular}{lll}
\toprule
\textbf{Model} & \textbf{MSE} &\\\midrule
\textbf{Latent Factor Models (LFM)} & &\\
\myindent Simple LFM~\cite{korenKDD2008} & 0.95 &\\
\myindent Experience-based LFM~\cite{mcauley2013} & 0.85 &\\
\myindent Text-based LFM~\cite{mcauleyrecsys2013} & 0.78 &\\
\textbf{Our Model: User SVR} & 0.60 &\\
\bottomrule
\end{tabular}
\vspace{-0.5em}
\caption{MSE comparison of models for predicting users' credibility rating
behavior with $10$-fold cross-validation. Improvements are statistically
significant with {\em P-value} < $0.0001$.}
\vspace{0.5em}
\label{tab:MSE1}
\begin{tabular}{lll}
\toprule
\textbf{Model} & \textbf{Only} & \textbf{Title}\\
& \textbf{Title} & \textbf{\& Text}\\
& \textbf{MSE} & \textbf{MSE}\\\midrule
\textbf{Language Model: SVR} & &\\
\myindent Language (Bias and Subjectivity) & 3.89 & 0.72\\
\myindent Explicit Topics & 1.74 & 1.74\\
\myindent Explicit + Latent Topics & 1.68 & 1.01\\
\myindent All Topics (Explicit + Latent) + Language & 1.57 & 0.61 \\\midrule

\textbf{News Source Features and Language Model: SVR} & & \\
\myindent News Source & 1.69 & 1.69\\
\myindent News Source + All Topics + Language & 0.91 & 0.46\\\midrule

\textbf{Aggregated Model: SVR} & & \\
\myindent Users + All Topics + Language + News Source & 0.43 & 0.41\\\midrule

\textbf{Our Model: CCRF+SVR} & & \\
\myindent User + All Topics + Language + News Source & 0.36 & 0.33\\
%User (Topic Preference) + All Topics + Language + News Source + User Features & 0.63\\
\bottomrule
\end{tabular}
\vspace{-0.5em}
\caption{MSE comparison of models for predicting aggregated article credibility rating with $10$-fold cross-validation. Improvements are statistically significant with {\em P-value} < $0.0001$.}
\vspace{-1em}
\label{tab:MSE}
%\end{minipage}
\end{table}

\begin{table}
\parbox{0.45\linewidth}{
\centering
\small
\begin{tabular}{lc}
\toprule
\textbf{Model} & \textbf{NDCG}\\
\midrule
Experience LFM~\cite{mcauley2013} & 0.80\\
PageRank & 0.83\\
CCRF & 0.86\\
\bottomrule
\end{tabular}
\vspace{-0.5em}
\caption{NDCG scores for ranking trustworthy sources.}
\label{tab:NDCGsources}
}
\hfill
\parbox{0.45\linewidth}{
\centering
\small
\begin{tabular}{lc}
\toprule
\textbf{Model}  & \textbf{NDCG}\\
\midrule
Experience LFM~\cite{mcauley2013} & 0.81\\
Member Ratings & 0.85\\
CCRF & 0.91\\
\bottomrule
\end{tabular}
\vspace{-0.5em}
\caption{NDCG scores for ranking expert users.}
\label{tab:NDCGmembers}
}
\vspace{-1.5em}
\end{table}

\begin{comment}
//all feautures- SVR - 0.84
//all feautures SVR + SVR - 0.29
explains why this is happening?
\end{comment}

%%%%%%%%%%%%%%%%%%%%%%%%%%%%%%%%%%

\subsection{Predicting User Ratings of Articles}

First we evaluate how good our model can predict the credibility ratings that users
assign to news articles using the \textit{Mean Squared Error} (MSE) between the prediction
and the actual rating. 
%Table~\ref{tab:evaldata} shows the evaluation data statistics.

\noindent{\bf Baselines}: We consider the following baselines for comparison:\\
{\bf 1.} {\em Latent Factor Recommendation Model} (LFM)~\cite{korenKDD2008}: LFM considers the tuple $\langle userId, itemId, rating \rangle$, and models each user and item as a vector of latent factors which are learned by minimizing the MSE between the rating and the product of the user-item latent factors. In our setting, each news article is considered an item and rating refers to the credibility rating assigned by a user to an article.\\
{\bf 2.} {\em Experience-based LFM}~\cite{mcauley2013}: This model incorporates {\em experience} of a user in rating an item in the LFM. The model builds on the hypothesis that users at similar levels of experience have similar rating behaviors which evolve with {\em time}. The model has an extra dimension: the {\em time} of rating an item which is not used in our SVR model. Note the analogy between the {\em experience} of a user in this model, and the notion of user {\em expertise} in the SVR model. However, these models ignore the text of the reviews. \\
{\bf 3.} {\em Text-based LFM}~\cite{mcauleyrecsys2013}: This model incorporates text in the LFM by combining the latent factors associated to items in LFM with latent topics in text from topic models like LDA.\\ %Unlike our model, which learns the latent topics separately, the baseline model learns the ratings and latent topics jointly.\\
{\bf 4.} {\em Support Vector Regression} (SVR)~\cite{drucker1996}: We train an SVR model $\text{SVR}_{u_k}$ for {\em each} user $u_k$ (refer to Section~\ref{subsec:SVR}) based on her reviews $\langle r_{j,k} \rangle$ with language and topic features $\langle F^L(r_{j,k}) \cup F^T(r_{j,k}) \rangle$, with the user's article ratings $\langle y_{j,k} \rangle$ as the response variable. We also incorporate the article language features and the topic features, as well as source-specific features to train the user model for this task. The other models ignore the stylistic features, and other fine-grained {\em user-item} interactions in the community.

Table~\ref{tab:MSE1} shows the MSE comparison between the different methods.
Our model (User SVR) achieved the lowest MSE and thus performed best.

%%%%%%%%%%%%%%%%%%%%%%%%%%%%%%%%%%
\vspace{-0.5em}
\subsection{Finding Credible Articles}

As a second part of the evaluation, we investigate the predictive power of different models in order to find credible news articles based on the
{\em aggregated ratings from all users}. The above LFM models, unaware of the {\em user cliques}, cannot be used directly for this task, as each news article has multiple reviews from different users which need to be aggregated. 
%to get an overall credibility rating for the article.
%to predict the overall credibility.
We find the \textit{Mean Squared Error} (MSE) between the estimated overall article rating, and the ground-truth article rating. We consider stories with {\em at least 3 ratings} about a news article.
% we managed to extract from their original sources. 
We compare the CCRF against the following baselines:\\
{\bf 1.} {\em Support Vector Regression} (SVR)~\cite{drucker1996}: We consider an SVR model with features on language (bag-of-all-words, subjectivity, bias etc.), topics (explicit tags as well as latent dimensions), and news-source-specific features. The language model uses all the lexicons derived and used in~\cite{recasens2013,mukherjee2014}. 
%The topic model is similar to the one used in~\cite{mukherjee2013WWW}. 
The source model also includes topic features in terms of the top topics covered by the source, and its topic-specific expertise for a subset of the topics.\\
{\bf 2.} {\em Aggregated Model} (SVR)~\cite{drucker1996}: As explained earlier, the user features cannot be directly used in the baseline model, which is agnostic of the user {\em cliques}. Therefore, we adopt a simple aggregation approach by taking the {\em average} rating of all the user ratings $\frac{\text{SVR}_{u_k}(d_j)}{|u_k|}$ for an article $d_j$ as a feature. 
Note that, in contrast to this simple average used here, our CCRF model learns the weights $\langle \alpha_u \rangle$ {\em per-user} to
combine their overall ratings for an article.

Table~\ref{tab:MSE} shows the MSE comparison of the different models.

\noindent{\bf MSE Comparison}: The first two models in Table~\ref{tab:MSE1} ignore the textual content of news articles, and reviews, and perform worse than the ones 
that incorporate full text. The text-based LFM considers title and text, and performs better than its predecessors. However, the User SVR model considers richer features and interactions, and attains $23\%$ MSE reduction over the best performing
LFM baselines.
%in the first task.

The baselines in Table~\ref{tab:MSE} show the model performance after incorporating different features in two different settings:
1) with news article {\em titles} only as text, and 
2) with titles and the {\em first few paragraphs} of an article.
The language model, especially the bias and subjectivity features, is less effective using only the article titles due to sparseness. 
On the other hand, using the entire article text may lead to very noisy features. 
So including the first few paragraphs of an article is the ``sweet spot''.
For this, we made an ad-hoc decision and included the first $1000$ characters
of each article.
With this setting, the language features made a substantial contribution to
reducing the MSE.
%{\tt GW: we may want to reconsider the wording here -- it should not sound like substantial tuning in order to get language features to work!!!!!!}\\

%Using language and topic features along with the {\em news source} specific features %attain a significant performance improvement over the news source features in isolation.
%%%GW: drop this, focus on primary message here

The aggregated SVR model further brings in the {\em user} features, and achieves
the lowest MSE among the baselines. 
This shows that a user-aware credibility model performs better than
user-independent ones.
Our CCRF model combines all features in a more sophisticated manner,
which results in $19.5\%$ MSE reduction over the most competitive baseline (aggregated SVR).
This is empirical evidence that the {\em joint} interactions between the different factors in a news community are indeed important to consider for identifying highly credible articles.
%\vspace{-0.5em}

%{\tt GW: shouldn't we have also NDCG scores for the top-k most credible articles? This would make this study perfectly analogous to the ones on trust and expertise!!!!!!}\\

%%%%%%%%%%%%%%%%%%%%%%%%%%%%%%%
\vspace{-0.5em}
\subsection{Finding Trustworthy Sources}
We shift the focus to two  use cases: 1) identifying the most trustworthy sources,
 and 2) identifying expert users in the community who can play the role of ``citizen journalists''.

Using the model of Section~\ref{sec:inference}, we rank all news sources in the community according to the learned $\langle \beta_{s_i} \rangle$ in Equation~\ref{eq.7}. The baseline is taken as the \emph{PageRank} scores of news sources in the Web graph. In the
experience-based LFM we can consider the sources to be users, and articles generated by them to be items. This allows us to obtain a ranking of the sources based on their 
%expertise
overall authority.
%{\tt GW: I would reserve "expertise" for individual users. Here it seems that the user expertise levels are aggregated by source. This leads to a notion similar to HITS-like authority IMO. May still have to discuss this!!!!!}
This is the second baseline against which we compare the CCRF.

We measure the quality of the ranked lists in terms of {\em NDCG} using the actual ranking of the news sources in the community as ground-truth.
%Kalervo Järvelin, Jaana Kekäläinen:
%Cumulated gain-based evaluation of IR techniques. ACM Trans. Inf. Syst. 20(4): 422-446 (2002)
NDCG gives geometrically decreasing weights to predictions at the various positions of the ranked list:

{\small
$NDCG_p = \frac{DCG_p}{IDCG_p}$ where
$DCG_p = rel_1 + \sum_{i=2}^p \frac{rel_i}{\log_2 i}$
}

Table \ref{tab:NDCGsources} shows the NDCG scores for the different methods.
%{\tt GW: to be continued !!!!!}\\

\vspace{-0.5em}
\subsection{Finding Expert Users}

Similar to news sources, we rank users according to the learned $\langle \alpha_{u_k} \rangle$ in Equation~\ref{eq.7}. The baseline is the average rating received by a user from other members in the community. We compute the NDCG score for the ranked lists of users
by our method.  We also compare against the ranked list of users from the experience-aware LFM~\cite{mcauley2013}. 
%another
%GW: ground-truth is not a baseline method, so LFM is the only baseline here
%baseline.
Table \ref{tab:NDCGmembers} shows the NDCG scores for different methods.
%{\tt GW: to be continued !!!!!}\\

%%%%%%%%%%%%%%%%%%%%%%%%%%%%%%%%%

\comm{
%\subsection{Qualitative Analysis}
\subsection{Qualitative Findings}
\label{subsec:qualanalysis}

\noindent{\bf Source Trustworthiness}: In general, 
science and technology websites (e.g., discovermagazine.com, nature.com, scientificamerican.com), investigative reporting and non-partisan sources (e.g., truthout.org, truthdig.com, cfr.org), book sites (e.g., nybooks.com, editorandpublisher.com), encyclopedia (e.g., Wikipedia)
and fact checking sites (e.g., factcheck.org) rank among the top trusted sources. 
Table~\ref{tab:sampletopics} shows the most and least trusted sources on four sample topics.
Overall, news sources are considered trustworthy with an average rating of
$3.46$ and variance of $0.15$.

%%%GW: overall, I would leave this out - it would be interesting for political opinion analysis, but it is very US-centric and susceptible to all kinds of subjective debates
%
\begin{comment}
%
Tables~\ref{tab:sampleview} and~\ref{tab:samplemedia} show the most and least trusted sources 
%for the most popular categories under the different facets.
%%%GW: not clear what categories and facets mean here
%I suppose all this refers to political viewpoints, if so we should say it clearly
for different political viewpoints.

Contents from sources with \emph{left} %(e.g., democracynow.org, truthdig.com)
and \emph{neutral} %(e.g., cfr.org, editorandpublisher.com)
viewpoints are more likely to be posted in the community than that from \emph{right} %(e.g., foxnews.com, rightwingnews.com)
sources. Contents from sources with \emph{national} and \emph{international} viewership are more likely to be posted due to their large audience base.
%Although, the scope of the source does not seem to have a strong effect on the credibility of the article.
%
\end{comment}

Table \ref{tab:samplemedia} shows the most and least trusted sources
for different media types.
Contents from \emph{blogs} %(e.g., juancole.com, dailykos.com)
are most likely to be posted followed by newspaper, magazine %(e.g., rollingstone.com, nybooks.com)
and other online sources. Contents from \emph{wire service, TV} and \emph{radio} are deemed the most trustworthy, although they have the least subscription, followed by \emph{magazines}. %are also very much trustworthy and they constitute a good proportion of the posted articles.

%%%GW: this paragraph is another candidate for dropping
\begin{comment}
Articles from the genres \emph{Comedy News} (e.g., huffingtonpost.com/ comedy), \emph{Investigative Report} (e.g., washingtonpost.com/investigations) and \emph{Fact Check} (e.g., politifact.org/truth-o-meter) are posted the least, but are deemed the most trustworthy. Articles from the genres \emph{Poll} and \emph{Editorial} are the least trustworthy; whereas articles from \emph{Opinion} and \emph{News Report} are posted the most.
\end{comment}

\begin{table}
\center
\scriptsize
\begin{tabular}{p{0.8cm}p{0.8cm}p{0.8cm}p{0.8cm}}
\toprule
\multicolumn{1}{c}{\textbf{Money - Politics}} & \multicolumn{1}{c}{\textbf{War in Iraq}} & \multicolumn{1}{c}{\textbf{Media - Politics}} & \multicolumn{1}{c}{\textbf{Green Technology}} \\\midrule
\multicolumn{1}{c}{\textbf{Most Trusted}} &  &  &  \\\midrule
rollingstone.com & nybooks.com & consortiumnews & discovermagazine.com \\
truthdig.com & consortiumnews & thenation.com & nature.com \\
democracynow.org & truthout.org & thedailyshow.com & scientificamerican.com \\
%thenation.com & juancole.com & youtube.com & guardian.co.uk \\
%youtube.com & mcclatchydc & newyorker.com & bbc.co.uk \\
%alternet.org & commondreams & theatlantic.com & gigaom.com \\
%gregpalast.com & prospect.org & salon.com & popsci.com \\
%motherjones.com & npr.org & gregpalast.com & technologyreview.com \\
%dailykos.com & harpers.org & factcheck.org & newscientist.com \\
\toprule
\multicolumn{1}{c}{\textbf{Least Trusted}} &  &  &  \\\midrule
firedoglake.com & crooksandliars & rushlimbaugh.com &  \\
suntimes.com & timesonline & rightwingnews.com &  \\
trueslant.com & suntimes.com & foxnews.com &  \\
%thehill.com & iht.com & firedoglake.com &  \\
%money.cnn.com & trueslant.com & msnbc.msn.com &  \\
%startribune.com & abcnews.go & crooksandliars.com &  \\
\bottomrule
\end{tabular}
\vspace{-1.5em}
\caption{Most and least trusted sources on sample topics.}
\vspace{-2em}
\label{tab:sampletopics}
\end{table}

\begin{comment}
\begin{table}
\center
	\scriptsize
	\begin{tabular}{p{1.7cm}p{1.9cm}p{2cm}p{1.8cm}}
		\toprule
		\textbf{Left}&\textbf{Right}&\textbf{Center}&\textbf{Neutral}\\\midrule
		\textbf{Most} & \textbf{Trusted}\\\midrule
		democracynow, truthdig.com, rollingstone.com & courant.com, opinionjournal.com, townhall.com & armedforces- journal.com, bostonreview.net & spiegel.de, cfr.org, editorandpublisher.com
		\\\midrule
		\textbf{Least} & \textbf{Trusted}\\\midrule
		crooksandliars, suntimes.com, washingtonmonthly.com & rightwingnews, foxnews.com, weeklystandard.com & sltrib.com, examiner.com, spectator.org & msnbc.msn.com, online.wsj.com, techcrunch.com
		\\\bottomrule
	\end{tabular}
	\caption{Most and least trusted sources with diff. viewpoints.}
	\label{tab:sampleview}
	\vspace{-1.5em}
\end{table}
\end{comment}

\begin{table}
\center
\scriptsize
\begin{tabular}{p{1.3cm}p{1.7cm}p{1.7cm}p{1.7cm}}
\toprule
\multicolumn{1}{c}{\textbf{Magazine}} & \multicolumn{1}{c}{\textbf{Online}} & \multicolumn{1}{c}{\textbf{Newspaper}} & \multicolumn{1}{c}{\textbf{Blog}} \\
\toprule
\multicolumn{1}{c}{\textbf{Most Trusted Sources}} &  &  &  \\\midrule
rollingstone.com & truthdig.com & nytimes.com & juancole.com \\
nybooks.com & cfr.org & nola.com & dailykos.com \\
thenation.com & consortiumnews & seattletimes & huffingtonpost \\
%nature.com & truthout.org & guardian.co.uk & nytimes.com \\
%newyorker.com & youtube.com & csmonitor.com & scholarsandrog. \\
%theatlantic.com & propublica.org & baltimoresun.com & thinkprogress \\
%motherjones.com & theatlantic.com & ft.com & wired.com \\
%economist.com & alternet.org & citypaper.com & globalvoices. \\
%tnr.com & salon.com & miamiherald.com & glenngreenwald \\
\midrule
\multicolumn{1}{c}{\textbf{Least Trusted Sources}} &  &  &  \\
\midrule
weeklystandard.com & investigativevoice & suntimes.com & rightwingnews\\
commentarymagazine & northbaltimore & nydailynews.com & firedoglake.com \\
nationalreview.com & hosted.ap.org & dailymail.co.uk & crooksandliars \\
%money.cnn.com & online.wsj.com & online.wsj.com & techcrunch.com \\
%forbes.com & wbal.com & telegraph.co.uk & buzzmachine \\
%baltimoremagazine.net & nationalreview & iht.com &  \\
%city-journal.org & baltimore & thehill.com &  \\
% & trueslant.com & chicagotribune.com &  \\
% & thehill.com & startribune.com &  \\
 \bottomrule
\end{tabular}
\caption{Most and least trusted sources on different types of media.}
\label{tab:samplemedia}
\vspace{-1em}
\end{table}

%%%%%%%%%%%%%%%%%%%%%%%%%%%%%
\noindent{\bf User Expertise}: We give some examples of the influence of user {\em viewpoint} and {\em expertise} on their reviews.\\
%{\tt GW: this part is very anecdotal and very susceptible to KDD-style critique that there is no methodological contribution here (in this part)!!!!! This is risky!!!!! Review and reconsider this part!!!!!!}

\example{
Viewpoint --- Low rating by non-expert: Sounds like the right wing nut jobs are learning how to use traditional leftist pincko tricks. It sucks its messy but it works.
}

In general, we find that community disagreement --- standard deviation in article credibility ratings by different members --- for different viewpoints are as follows: Right ($0.80$) > Left($0.78$) > Center($0.65$) > Neutral ($0.63$).

\textit{Expertise}: Following example shows an expert in nuclear energy \textit{downvoting} another user's rating on nuclear radiation.

\example{
Non-expert: Interesting opinion about the health risks of nuclear radiation, from a physicist at the University of Oxford. He makes some reasonable points based on factual evidence, ...\\
Low rating by expert to above review: Is it fair to assume that you have no background in biology or anything medical? While this story is definitely very important, it contains enough inaccurate and/or misleading statements that it should be interpreted with great caution.
}

\textit{Bias}: An article on {\scriptsize Racial-tension-simmers-on-Marthas-Vineyard-as-Barack-Obama-arrives} gets an overall low credibility rating. The first example shows an expert objectively evaluating the article, whereas the second one shows a non-expert's biased review.

\example{
Expert: This is an attempt at race-baiting. Why is Obama connected with this incident at all. I am sure the Brazilian laborers were there when George Bush was in office, as well.
%This is just wrong. the caption states ``racial tensions simmers" not immigration tensions. This is complete hogwash!!
}

\example{
Non-expert: Great piece because it highlights the hypocrisy that the power elites in this country have lead us into. Obama worried about what clams to suck up while I am worried about my family getting fed.
%I no longer tell my kids that hard work and merit are what counts. My son can't get a job making a few bucks doing anything because illegals have all the entry positions. %
}
}

\begin{table}
\centering
\scriptsize
 \begin{tabular}{ll}
 \toprule
 \textbf{Factors} & \textbf{Corr.}\\\midrule
 \textbf{a) }{Stylistic Indicators} Vs. Article Credibility Rating & \\
 \myindent Insightful (Is it well reasoned? thoughtful?) & 0.77\\
 \myindent Fairness (Is it impartial? or biased?) & 0.75\\
 \myindent Style (Is this story clear? concise? well-written?) & 0.65\\
 \myindent Responsibility (Are claims valid, ethical, unbiased?) & 0.72\\
 \myindent Balance (Does this story represent diverse viewpoints?) & 0.49\\\midrule
 \textbf{b) }Influence of Politics Vs. Disagreement & 0.11\\
 \textbf{c) }Expertise (Moderate, High) Vs. Disagreement & -0.10, -0.31\\
 {Interactions} & \\
 \myindent \textbf{d) } User Expertise Vs. User-User Rating & 0.40\\
 \myindent \textbf{e) } Source Trustworthiness Vs. Article Credibility Rating & 0.47\\
 \myindent \textbf{f) } User Expertise Vs. MSE in Article Rating Prediction & -0.29\\
 \bottomrule
 \end{tabular}
 \vspace{-0.5em}
 \caption{Pearson's product-moment correlation between various factors (with {\em P-value} $< 0.0001$ for each test).}
 \label{tab:corAspects}
  \vspace{-1.3em}
\end{table}

%\subsection{Summarizing Findings}
%%%GW: "summarizing" diminishes the value of this part

\subsection{Discussion}
\label{subsec:discussions}

%{\tt GW: wirte this in a sequence of ca. 3 paragraphs, not in an itemized list - itemization emphasizes summary, whereas paragraphs suggested true discussion with added-value insight !!!!!}

\noindent{\bf Hypothesis Testing}: We test various hypotheses under the influence of the feature groups using explicit labels, and ratings available in the NewsTrust community. A summary of the tests is presented in Table~\ref{tab:corAspects} showing a {\em moderate} correlation between various factors which are put together in the CCRF to have a {\em strong} indicator for information credibility.

%We present our experimental findings from Tables~\ref{tab:MSE1}-~\ref{tab:corAspects}.

\noindent{\bf Language}: The stylistic features (factor (a) in Table~\ref{tab:corAspects}) like \textit{assertives, hedges, implicatives, factives, discourse} and \textit{affective} play a significant role in the credibility detection of news, in conjunction with other language features like {\em topics}.
%%%GW: the table is very hard to interpret, as there is no detail given
%%%also: this seems to refer to explicit labels from NewsTrust, not to our model features ???
%%%I'm afraid the typical reader=reviewer will not get much out of this table
%Table~\ref{tab:corAspects} shows the correlation of various explicit stylistic indicators %rated by users with the overall article credibility rating.

\noindent{\bf Topics}: Topics are an important indicator for news credibility. We measured the influence of the {\em Politics} tag on other topics by their co-occurrence
frequency in the explicit tag sets over all the news articles. 
%This allows us to obtain the overall influence of politics on any article by averaging the %political association of each of its explicit topic tags. 
We found significant influence of Politics on all topics, with an average measure of association of $54\%$ to any topic, and $62\%$ for the overall news article.
%%%GW: the numbers are not really informative, as there is no comparison to other forums/media
The community gets polarized due to different perspectives on topical aspects of news. 
A moderate correlation (factor (b) in Table~\ref{tab:corAspects}) indicates a weak trend of disagreement, measured by the standard deviation in article credibility rating among users, increasing with its political content. In general, we find that community disagreement for different viewpoints are as follows: Right ($0.80$) > Left($0.78$) > Center($0.65$) > Neutral ($0.63$).

%{\tt GW: better trim this par -- it sounds a lot like very specific observations about NewsTrust!!!!!}\\

\noindent{\bf Users}: User engagement features are strong indicators of expertise.
%Users in the community have their own perspectives and expertise on various topics. Modeling user-specific topic perspectives and expertise helps to capture their user judgments on article credibility.
Although credibility is ultimately subjective, experts show moderate agreement (factor (c) in Table~\ref{tab:corAspects}) on highly credible
news. There is a moderate correlation (factor (d) in Table~\ref{tab:corAspects}) between feedback received by a user on his
ratings from community, and his expertise.

\noindent{\bf Sources}: Various traits of a news source like viewpoint, format and topic expertise are strong indicators of trustworthiness.
In general,
science and technology websites (e.g., discovermagazine.com, nature.com, scientificamerican.com), investigative reporting and non-partisan sources (e.g., truthout.org, truthdig.com, cfr.org), book sites (e.g., nybooks.com, editorandpublisher.com), encyclopedia (e.g., Wikipedia)
and fact checking sites (e.g., factcheck.org) rank among the top trusted sources.
Table~\ref{tab:sampletopics} shows the most and least trusted sources on four sample topics.
Overall, news sources are considered trustworthy with an average rating of $3.46$ and variance of $0.15$.
Tables \ref{tab:sampleview} and \ref{tab:samplemedia} show the most and least trusted sources on different viewpoints and media types respectively.
Contents from \emph{blogs} %(e.g., juancole.com, dailykos.com)
are most likely to be posted followed by newspaper, magazine %(e.g., rollingstone.com, nybooks.com)
and other online sources. Contents from \emph{wire service, TV} and \emph{radio} are deemed the most trustworthy, although they have the least subscription, followed by \emph{magazines}. %are also very much trustworthy and they constitute a good proportion of the posted articles.

\begin{table}
\center
\scriptsize
\begin{tabular}{p{0.8cm}p{0.8cm}p{0.8cm}p{0.8cm}}
\toprule
\multicolumn{1}{c}{\textbf{Money - Politics}} & \multicolumn{1}{c}{\textbf{War in Iraq}} & \multicolumn{1}{c}{\textbf{Media - Politics}} & \multicolumn{1}{c}{\textbf{Green Technology}} \\\midrule
\multicolumn{1}{c}{\textbf{Most Trusted}} &  &  &  \\\midrule
rollingstone.com & nybooks.com & consortiumnews & discovermagazine.com \\
truthdig.com & consortiumnews & thenation.com & nature.com \\
democracynow.org & truthout.org & thedailyshow.com & scientificamerican.com \\
%thenation.com & juancole.com & youtube.com & guardian.co.uk \\
%youtube.com & mcclatchydc & newyorker.com & bbc.co.uk \\
%alternet.org & commondreams & theatlantic.com & gigaom.com \\
%gregpalast.com & prospect.org & salon.com & popsci.com \\
%motherjones.com & npr.org & gregpalast.com & technologyreview.com \\
%dailykos.com & harpers.org & factcheck.org & newscientist.com \\
\toprule
\multicolumn{1}{c}{\textbf{Least Trusted}} &  &  &  \\\midrule
firedoglake.com & crooksandliars & rushlimbaugh.com &  \\
suntimes.com & timesonline & rightwingnews.com &  \\
trueslant.com & suntimes.com & foxnews.com &  \\
%thehill.com & iht.com & firedoglake.com &  \\
%money.cnn.com & trueslant.com & msnbc.msn.com &  \\
%startribune.com & abcnews.go & crooksandliars.com &  \\
\bottomrule
\end{tabular}
\vspace{-0.5em}
\caption{Most and least trusted sources on sample topics.}
\label{tab:sampletopics}
\vspace{-1em}
\end{table}

\begin{table}
\center
	\scriptsize
	\begin{tabular}{p{2cm}p{2.1cm}p{1.7cm}p{1.65cm}}
		\toprule
		\textbf{Left}&\textbf{Right}&\textbf{Center}&\textbf{Neutral}\\\midrule
		\textbf{Most} & \textbf{Trusted}\\\midrule
		democracynow, truthdig.com, rollingstone.com & courant.com, opinionjournal.com, townhall.com & armedforces- journal.com, bostonreview.net & spiegel.de,cfr.org, editorandpublisher.com
		\\\midrule
		\textbf{Least} & \textbf{Trusted}\\\midrule
		crooksandliars, suntimes.com, washingtonmonthly.com & rightwingnews, foxnews.com, weeklystandard.com & sltrib.com, examiner.com, spectator.org & msnbc.msn.com, online.wsj.com, techcrunch.com
		\\\bottomrule
	\end{tabular}
	\vspace{-0.5em}
	\caption{Most and least trusted sources with different viewpoints.}
	\label{tab:sampleview}
	\vspace{-1em}
\end{table}

\begin{table}
\center
\scriptsize
\begin{tabular}{p{1.3cm}p{1.7cm}p{1.7cm}p{1.5cm}}
\toprule
\multicolumn{1}{c}{\textbf{Magazine}} & \multicolumn{1}{c}{\textbf{Online}} & \multicolumn{1}{c}{\textbf{Newspaper}} & \multicolumn{1}{c}{\textbf{Blog}} \\
\toprule
\multicolumn{1}{c}{\textbf{Most Trusted Sources}} &  &  &  \\\midrule
rollingstone.com & truthdig.com & nytimes.com & juancole.com \\
nybooks.com & cfr.org & nola.com & dailykos.com \\
thenation.com & consortiumnews & seattletimes & huffingtonpost \\
%nature.com & truthout.org & guardian.co.uk & nytimes.com \\
%newyorker.com & youtube.com & csmonitor.com & scholarsandrog. \\
%theatlantic.com & propublica.org & baltimoresun.com & thinkprogress \\
%motherjones.com & theatlantic.com & ft.com & wired.com \\
%economist.com & alternet.org & citypaper.com & globalvoices. \\
%tnr.com & salon.com & miamiherald.com & glenngreenwald \\
\midrule
\multicolumn{1}{c}{\textbf{Least Trusted Sources}} &  &  &  \\
\midrule
weeklystandard.com & investigativevoice & suntimes.com & rightwingnews\\
commentarymagazine & northbaltimore & nydailynews.com & firedoglake.com \\
nationalreview.com & hosted.ap.org & dailymail.co.uk & crooksandliars \\
%money.cnn.com & online.wsj.com & online.wsj.com & techcrunch.com \\
%forbes.com & wbal.com & telegraph.co.uk & buzzmachine \\
%baltimoremagazine.net & nationalreview & iht.com &  \\
%city-journal.org & baltimore & thehill.com &  \\
% & trueslant.com & chicagotribune.com &  \\
% & thehill.com & startribune.com &  \\
 \bottomrule
\end{tabular}
\caption{Most and least trusted sources on different types of media.}
\label{tab:samplemedia}
\vspace{-1em}
\end{table}

\noindent{\bf Interactions}: In principle, there is a moderate correlation between \emph{trustworthy} sources generating \emph{credible} articles (factor (e) in Table~\ref{tab:corAspects}) identified by \emph{expert} users (factor (f) in Table~\ref{tab:corAspects}). A negative sign of correlation indicates decrease in disagreement or MSE with increase in expertise.
In a news community, we can observe {\em moderate} signals of interaction between various factors that characterize users, articles, and sources. Our CCRF model brings all these features together to build a {\em strong} signal for news credibility.

\vspace{-0.5em}

\section{Related Work}
\label{sec:related-work}

\noindent \textbf{Rating prediction in online communities:}
Collaborative filtering based approaches \cite{korenKDD2008} for rating
prediction exploit user and item similarities by latent factors.
\cite{mcauley2013} further studies the temporal evolution of users and their
rating behavior in this framework. Recent works~\cite{mcauleyrecsys2013,
mukherjee2014JAST} also tap into user review texts to generate user-specific
ratings of reviews. Other papers have studied temporal issues for anomaly detection~\cite{Gunnemann2014}.% and linguistic norms~\cite{DanescuWWW2013}. 
Prior work that tapped user review texts focused on other
issues. Sentiment analysis over reviews aimed to learn latent topics
~\cite{linCIKM2009}, latent aspects and their ratings~\cite{wang2011}, and
user-user interactions~\cite{West-etal:2014}. Our model unifies several
dimensions to jointly study the role of language, users, topics, and
interactions for information credibility.

\noindent \textbf{Information credibility in social media}: \cite{castillo2011} analyzes micro-blog postings in Twitter related to trending topics, and classifies them as credible or not based on features from user posting and re-posting behavior.
%They state that several factors influence the credibility of information in social media like ``the reactions that certain topics generate and the emotion conveyed by users discussing the topic: e.g. if they use opinion expressions representing positive or negative sentiments about the topic" and the level of certainty of the users and their characteristics in propagating that information.
%
\cite{kang2012} focuses on credibility of users, harnessing the dynamics of
information flow in the underlying social graph and tweet content.
\cite{canini2011} analyzes both topical content of information sources and
social network structure 
to find credible information sources in
social networks. Information credibility in tweets 
%corresponding to fourteen high impact news events of 2011 
has been studied in \cite{agupta2012}.  \cite{VZRPCIKM12} conducts a {\em user study} to analyze various factors like contrasting viewpoints and expertise affecting the truthfulness of controversial claims.
However, none of these prior works analyze the interplay between \textit{sources, language, topics,} and \textit{users}.
% that influences information credibility.

The works closest to our problem and approach are~\cite{vg2011,mukherjee2014}. 
\cite{vg2011} presents an algorithm for propagating trust scores in a heterogeneous network of claims, sources, and documents. \cite{mukherjee2014} proposes a method to jointly learn user trustworthiness, statement credibility, and language objectivity in online health communities. However, these works do not analyze the role of \textit{topics, language bias, user perspective, expertise}, and fine-grained interactions in community.

\noindent \textbf{Bias in social communities and media}: The use of biased language in Wikipedia and similar collaborative communities has been studied in~\cite{greene2009,recasens2013}. Even more broadly, the task of characterizing subjective language has been addressed, among others, in \cite{wiebe2005,lin2011}.
The influence of different kinds of bias in online user ratings has been studied in~\cite{sloanreview,fang2014}. \cite{fang2014} proposes an approach to handle users who might be subjectively different or strategically dishonest.

%%%GW: I would leave the following par out -- this gets more and more into media psychology and sociology, while at the same time being relatively old work whose applicability to modern Internet media may be debatable (everything older than 5 years seems weakly applicable at best)
\begin{comment}
\cite{chomsky1988} discuss a propaganda model of five editorially distorting filters applied to news reporting in mass media
%: ``size, ownership, and profit orientation; the advertising license to do business; sourcing mass media news; flak and the enforcers; anti-communism''.
``Does media bias affect voting?'' is studied in~\cite{dellavigna2007}
%by looking at the entry of Fox News in cable markets and its impact on voting.
where they find a significant influence of Fox News on the vote share in Presidential elections between 1996 and 2000, with the Republicans gaining 0.4 to 0.7 percentage points in the towns which broadcast Fox News. A very basic, beginning course in detecting the role of news media in spreading bias and propaganda is presented in~\cite{paul2006}.
\end{comment}

\noindent \textbf{Citizen journalism}: \cite{wemedia} defines citizen journalism as ``the act of a citizen or group of citizens playing an active role in the process of collecting, reporting, analyzing and dissemination of news and information to provide independent, reliable, accurate, wide-ranging and relevant information that a democracy requires.''
\cite{allan2007} focuses on user activities like blogging in community news websites.
%\cite{lewis2010} define citizen journalists as those who comment on stories, respond to polls, and submit video, audio and text to traditional media companies. \cite{garbett2014} present a set of design implications for building systems that support interaction between citizen and professional journalists.
Although the potential of citizen journalism is greatly highlighted in the recent Arab Spring~\cite{howard2011}, misinformation can be quite dangerous when relying on users as news sources (e.g., the reporting of the Boston
Bombings in 2013~\cite{boston}).
\vspace{-0.5em}

\section{Conclusions}

In this work, we analyzed the effect of different factors like \textit{language, topics} and \textit{perspectives} on the credibility rating of articles in a news community.
These factors and their mutual interactions are the features of
a novel model for jointly capturing \textit{credibility} of news articles, \textit{trustworthiness} of news sources and \textit{expertise} of users. 
%, who can perform the role of citizen journalists in a news community.
%Topics, users and sources form a major component of this work, where we %exploit the preference, expertise and viewpoint of the entities to establish %credibility of the content. 
From an application perspective, we demonstrated that our method can reliably identify credible articles, trustworthy sources and expert users in the community.

As future work, we plan to model and analyze the \textit{temporal} evolution of the factors associated with each of the components in our model. We have a strong intuition that time has a significant influence on the trustworthiness of sources and credibility of news.
%\nocite{*}
\vspace{-0.5em}
\bibliographystyle{abbrv}
\bibliography{peopleonmedia}

\end{document}